\documentclass[journal]{IEEEtran}
\usepackage{amsmath}
\usepackage{algorithm}
\usepackage{algpseudocode} % algorithmicx
\usepackage{caption}
\usepackage{cite}
\usepackage{tikz}
\usetikzlibrary{
    positioning,
    arrows.meta,
    shapes.geometric
}
\usepackage{amssymb}
\usepackage{graphicx}
\usepackage{pgfplots}
\usepackage{pgfplotstable}
\pgfplotsset{compat=1.18}
\usepackage{tabularx}
\usepackage{booktabs}
\usepackage{makecell}
\usepackage{array}
\usepackage{ragged2e}
\usepackage{stfloats}
\usepackage{hyperref}
\hypersetup{hidelinks}
\usepackage{enumitem}

\newcolumntype{L}{>{\raggedright\arraybackslash}X}
\newcolumntype{C}{>{\centering\arraybackslash}X}

\ifCLASSINFOpdf

\else

\fi

\begin{document}

\title{Agentic Workflow Using RBA$_\theta$ for Event Prediction}

\author{
Purbak~Sengupta$^{\dagger}$,
Sambeet~Mishra$^{\ddagger}$,
Sonal~Shreya$^{\dagger}$\\[0.5em]
$^{\dagger}$Department of Electrical and Computer Engineering, Aarhus University, Denmark\\
E-mail: sen.purbak@gmail.com, sshreya@ece.au.dk\\[0.3em]
$^{\ddagger}$Department of Electrical, IT and Cybernetics, University of South-Eastern Norway, Norway\\
E-mail: sambeet.mishra@usn.no
}

\maketitle

\begin{abstract}
Wind power ramp events are difficult to forecast due to strong variability,
multi-scale dynamics, and site-specific meteorological effects. This paper proposes an \emph{event-first, frequency-aware} forecasting paradigm that directly predicts ramp events and reconstructs the power trajectory
thereafter, rather than inferring events from dense forecasts. The framework is built on an enhanced Ramping Behaviour Analysis (RBA$_\theta$) method's event representation and progressively integrates statistical, machine-learning, and deep-learning models. Traditional forecasting models with post-hoc event extraction provides a strong interpretable baseline but exhibits limited generalisation across sites. Direct event prediction using Random Forests improves robustness over survival-based formulations, motivating fully event-aware modelling. To capture the multi-scale nature of wind ramps, we introduce an event-first deep architecture that integrates wavelet-based frequency decomposition, temporal excitation features, and adaptive feature selection. The resulting sequence models enable stable long-horizon event prediction, physically consistent trajectory reconstruction, and zero-shot transfer to previously unseen wind farms. Empirical analysis shows that ramp magnitude and duration are governed by distinct mid-frequency bands, allowing accurate signal reconstruction from sparse event forecasts. An agentic forecasting layer is proposed, in which specialised workflows
are selected dynamically based on operational context.
Together, the framework demonstrates that event-first, frequency-aware forecasting
provides a transferable and operationally aligned alternative to trajectory-first wind-power prediction.

\end{abstract}


\IEEEpeerreviewmaketitle

\section{Introduction}

\IEEEPARstart{T}{he} rapid global transition toward renewable energy has positioned wind power as a central pillar of modern electricity systems. Installed wind capacity surpassed one terawatt by 2023, with continued expansion driven by cost reductions, offshore deployment, technological maturity, and policy commitments toward low-carbon energy systems. As wind power evolves from a supplementary resource to a dominant source of electricity, its intrinsic variability arising from atmospheric processes spanning multiple temporal and spatial scales has emerged as a critical challenge for grid stability, reserve planning, and market operations.

Accurate wind power forecasting is therefore essential for operational decision-making, including scheduling, dispatch, maintenance, and energy trading. However, wind power generation exhibits pronounced nonlinearity, stochasticity, and nonstationarity, leading to large forecast uncertainty, particularly during periods of rapid power variation. Wind power ramp events, characterized by abrupt changes in generation, are especially consequential: they account for a disproportionate share of forecast error and operational risk, yet remain among the most difficult phenomena to predict reliably \cite{sawant, gallego}. Improving forecasting performance during such rare but impactful events is thus a central requirement for renewable-dominated power systems.

\subsection{Background}
\label{subsec:background}

According to the \href{https://www.irena.org/-/media/Files/IRENA/Agency/Publication/2024/Mar/IRENA_RE_Capacity_Statistics_2024.pdf}{International Renewable Energy Agency}, global installed wind capacity surpassed 1~TW in 2023, contributing close to 10\% of total electricity generation. As this share continues to increase, system operators face growing challenges related to variability, stability, and real-time balancing. Among the various manifestations of wind variability, ramp events represent the most critical challenge for grid operations. These events correspond to abrupt increases or decreases in power output that can trigger reserve activation, curtailment, or emergency trading, leading to substantial economic and operational consequences. Beyond financial impact, ramp events also influence turbine fatigue, grid-code compliance, and the coordination of multiple renewable resources. Empirical studies have shown that ramp events contribute disproportionately to forecasting error and grid stress \cite{gallego}. At short time horizons below six hours, these effects are dominated by chaotic micro and mesoscale dynamics, limiting the predictive capability of purely physical models. Despite advances in numerical weather prediction, statistical modeling, and deep learning, most existing forecasting approaches remain fundamentally trajectory-centric. These methods optimize global error metrics such as Root mean squared error (RMSE) or Mean absolute error (MAE) over continuous time series, implicitly assuming that improved average trajectory accuracy translates into better operational performance. In practice, however, grid operators and control systems respond primarily to discrete events such as ramp onset, duration, magnitude, and regime transitions rather than to small pointwise deviations during stable periods. As a result, strong global accuracy can mask systematic failures during operationally critical intervals, particularly when forecasts misalign ramp timing or underestimate ramp intensity.

This mismatch reflects a structural limitation of trajectory-first pipelines, in which event detection is applied post hoc to predicted power trajectories. Such pipelines require uniformly accurate predictions across all timesteps, even though only a sparse subset of the signal governs operational decisions. Moreover, trajectory-based representations are highly site-specific, reducing model transferability across wind farms with differing turbine configurations, terrain, and atmospheric conditions. These limitations motivate a shift away from dense time-domain regression toward representations that explicitly encode the semantic structure of wind power dynamics.

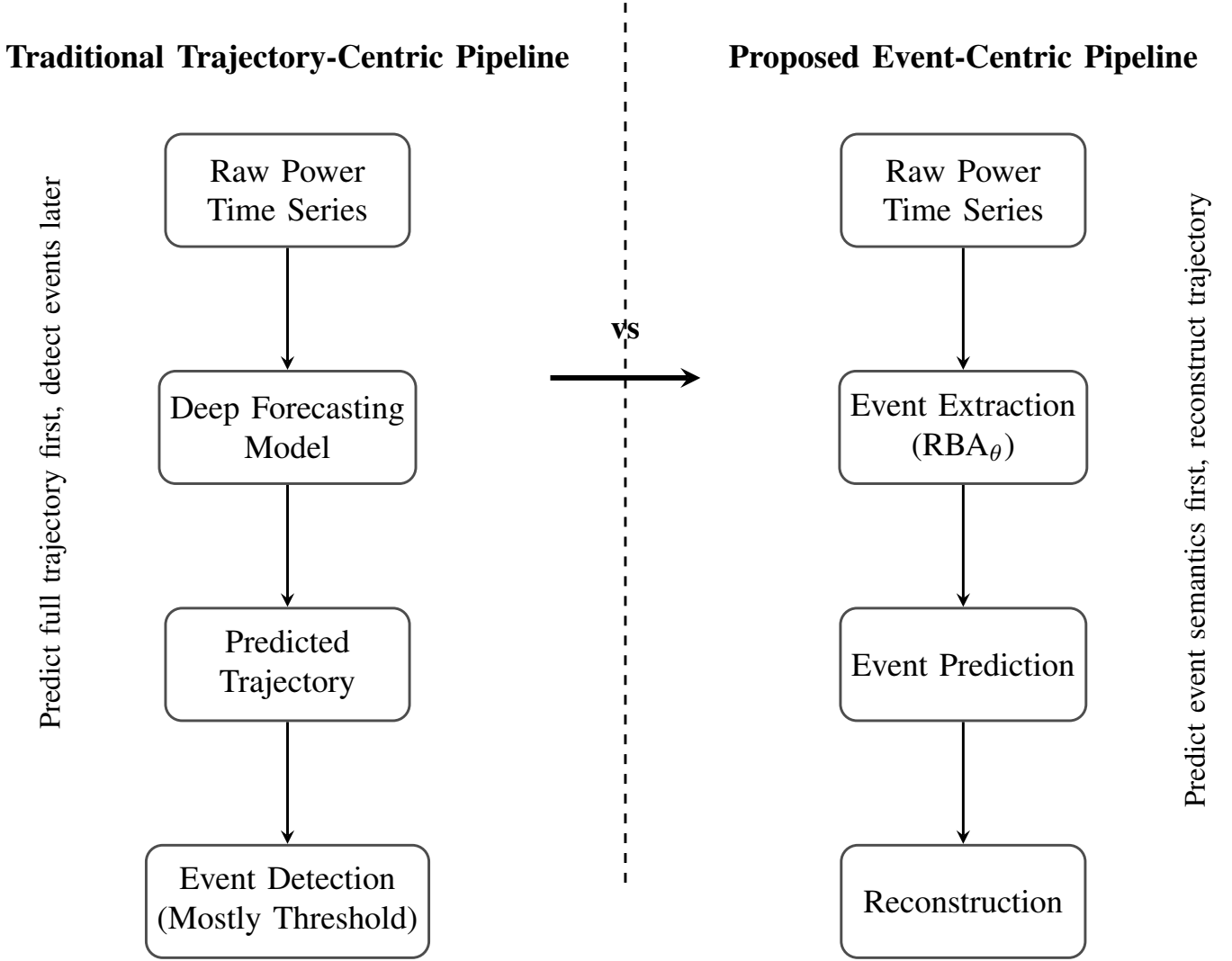
\begin{figure*}[t]
\centering
\resizebox{\textwidth}{!}{%
\begin{tikzpicture}[
    node distance=1.3cm,
    box/.style={
        rectangle,
        rounded corners=5pt,
        draw=black!70,
        fill=white,
        thick,
        minimum height=1.2cm,
        minimum width=2.6cm,
        align=center
    },
    arrow/.style={->, thick, >=stealth}
]

%%%%%%%%%%%%%%%%%%%%%%%%%%%%%%%%%%%%%
% GLOBAL X POSITIONS
%%%%%%%%%%%%%%%%%%%%%%%%%%%%%%%%%%%%%
\def\xleft{-3.6}
\def\xright{3.6}
\def\xcenter{0}

%%%%%%%%%%%%%%%%%%%%%%%%%%%%%%%%%%%%%
% LEFT PIPELINE
%%%%%%%%%%%%%%%%%%%%%%%%%%%%%%%%%%%%%
\node[align=center, font=\bfseries] (classicTitle)
    at (\xleft,3.2)
    {Traditional Trajectory-Centric Pipeline};

\node[box] (rawTS) at (\xleft,1.8)
    {Raw Power\\Time Series};

\node[box] (model) [below=of rawTS]
    {Deep Forecasting\\Model};

\node[box] (forecastTS) [below=of model]
    {Predicted\\Trajectory};

\node[box] (detect) [below=of forecastTS]
    {Event Detection\\(Mostly Threshold)};

\draw[arrow] (rawTS) -- (model);
\draw[arrow] (model) -- (forecastTS);
\draw[arrow] (forecastTS) -- (detect);

\node[rotate=90] at (\xleft-2.5,-1.0)
    {\small Predict full trajectory first, detect events later};

%%%%%%%%%%%%%%%%%%%%%%%%%%%%%%%%%%%%%
% RIGHT PIPELINE
%%%%%%%%%%%%%%%%%%%%%%%%%%%%%%%%%%%%%
\node[align=center, font=\bfseries] (eventTitle)
    at (\xright,3.2)
    {Proposed Event-Centric Pipeline};

\node[box] (rawTS2) at (\xright,1.8)
    {Raw Power\\Time Series};

\node[box] (extract2) [below=of rawTS2]
    {Event Extraction\\(RBA$_\theta$)};

\node[box] (predict2) [below=of extract2]
    {Event Prediction};

\node[box] (reconstruct2) [below=of predict2]
    {Reconstruction};

\draw[arrow] (rawTS2) -- (extract2);
\draw[arrow] (extract2) -- (predict2);
\draw[arrow] (predict2) -- (reconstruct2);

\node[rotate=90] at (\xright+2.5,-1.5)
    {\small Predict event semantics first, reconstruct trajectory};

%%%%%%%%%%%%%%%%%%%%%%%%%%%%%%%%%%%%%
% CENTER DIVIDER
%%%%%%%%%%%%%%%%%%%%%%%%%%%%%%%%%%%%%
\draw[dashed, thick]
    (\xcenter,3.8) -- (\xcenter,-5.6);

\draw[arrow, ultra thick]
    (\xcenter-0.8,-0.2) -- (\xcenter+0.8,-0.2);

\node at (\xcenter,0.3) {\bfseries vs};

\end{tikzpicture}}
\caption{Conceptual comparison between the traditional trajectory-first forecasting
pipeline and the proposed event-centric paradigm. The classical approach forecasts
the full time series and detects events afterward, whereas the proposed system
predicts event semantics first and reconstructs the trajectory, improving
interpretability, transferability, and operational relevance}
\label{fig:pipeline_comparison}
\end{figure*}
%==================================================

Wind power variability is inherently multi-scale. Atmospheric processes ranging from sub-minute turbulence and wake interactions to mesoscale fronts and synoptic weather systems imprint distinct spectral signatures on power time series \cite{orlanski}. Empirical analyses show that ramp dynamics are concentrated within specific frequency bands, while low-frequency components govern persistence and high-frequency components largely reflect turbulence and noise. This structure suggests that ramp events are not arbitrary irregularities but organized, scale-dependent phenomena. Forecasting models that operate solely in the time domain, without explicit multi-resolution or frequency-aware representations, are therefore intrinsically limited in their ability to capture the mechanisms underlying extreme transitions.

Recent deep learning architectures have improved nonlinear modeling capacity through convolutional, recurrent, and attention-based components. Nevertheless, most remain optimized for average predictive accuracy, leading to reduced sensitivity to rare but operationally dominant events. Predictive uncertainty typically increases during periods of rapid variation precisely when accurate forecasts are most critical and performance often degrades under distributional shifts across sites. These challenges highlight the need for forecasting paradigms that explicitly prioritize rare-event modeling, robustness, and transferability.

This work adopts an event-centric perspective in which forecasting is reframed around discrete ramp semantics rather than continuous trajectories. Building on the enhanced RBA$_\theta$ framework \cite{my_rba}, which provides deterministic and physically interpretable extraction of ramp events, the proposed approach treats event characteristics like onset, magnitude, duration, steepness, and type as primary predictive targets. Unlike prior studies that employ event descriptors for static thresholding based post-hoc analysis \cite{rba_sambeet}, the proposed paradigm inverts the conventional workflow by forecasting event structure first and reconstructing the continuous trajectory from these predictions.

Event-level representations offer two key advantages. First, they provide a naturally compressed description of wind dynamics, focusing learning capacity on the small subset of phenomena that dominate operational relevance. Second, if ramp semantics exhibit partially shared structure across sites, event-based abstractions are more likely to transfer than raw trajectories, enhancing generalization across heterogeneous wind farms. The RBA$_\theta$ mechanism yields stable and reproducible event labels, forming a consistent representation space for event-aware learning.

\begin{figure*}[t]
\centering
\includegraphics[width=\textwidth]{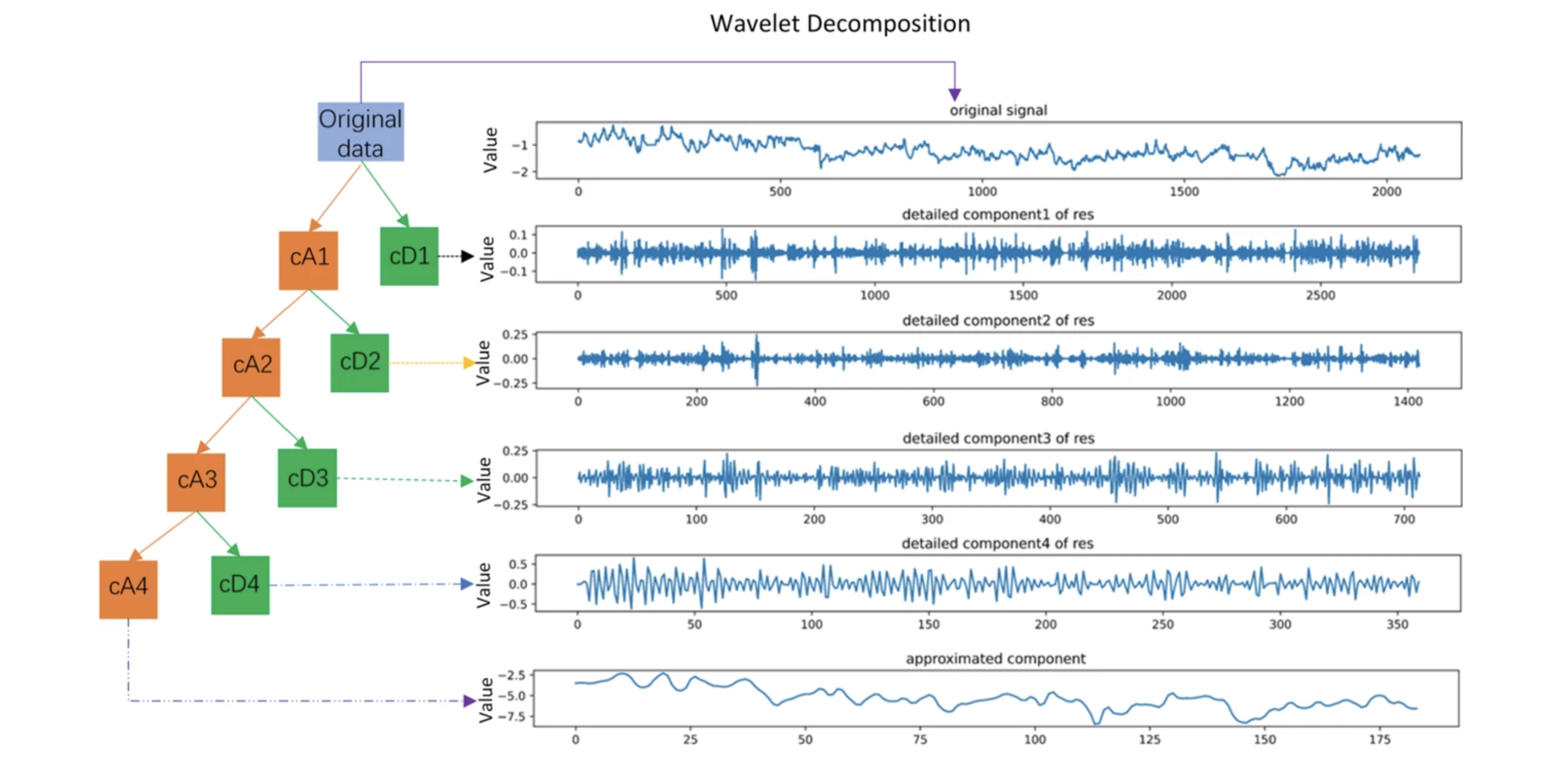}
\caption{Wavelet-based multi-band decomposition of the input signal, showing
approximation ($cA_i$) and detail ($cD_i$) components across scales \cite{wang_wavelet}}
\label{fig:wavelet}
\end{figure*}

To capture the multi-scale nature of wind dynamics, the proposed framework integrates frequency-aware modeling via wavelet-based signal decomposition \cite{singh, wang_wavelet}. Discrete wavelet transforms separate low-frequency persistence from intermediate-scale ramp dynamics and high-frequency turbulence, enabling models to focus on physically meaningful patterns while attenuating noise as shown in Fig~\ref{fig:wavelet}. This decomposition supports more stable learning of event characteristics compared to time-domain modeling alone.

%==================== FIGURE 3 ====================
%==================================================

However, no single forecasting model can robustly address the diverse conditions encountered in renewable energy systems, including varying horizons, uncertainty regimes, data availability, and distributional shifts. This motivates an agentic forecasting paradigm in which multiple specialized workflows are coordinated through a decision layer that adapts to contextual signals such as forecast horizon, regime volatility, uncertainty level, and transfer requirements. Agentic and mixture-of-experts frameworks have demonstrated improved robustness and interpretability across complex forecasting tasks \cite{jacob, hamilton, gneiting}. Within this paradigm, challenging periods such as ramp-dominated or uncertainty-heavy regimes can be routed to workflows explicitly designed for rare-event prediction, while simpler models handle stable conditions. By shifting predictive focus from dense trajectories to sparse, semantically meaningful events, the proposed framework aligns forecasting objectives with operational decision-making and improves robustness under uncertainty. Wind power serves as a challenging and practically relevant test case, but the proposed agentic, event-centric paradigm is broadly applicable to forecasting problems in which rare, high-impact events dominate operational risk.

%==================== FIGURE 4 ====================
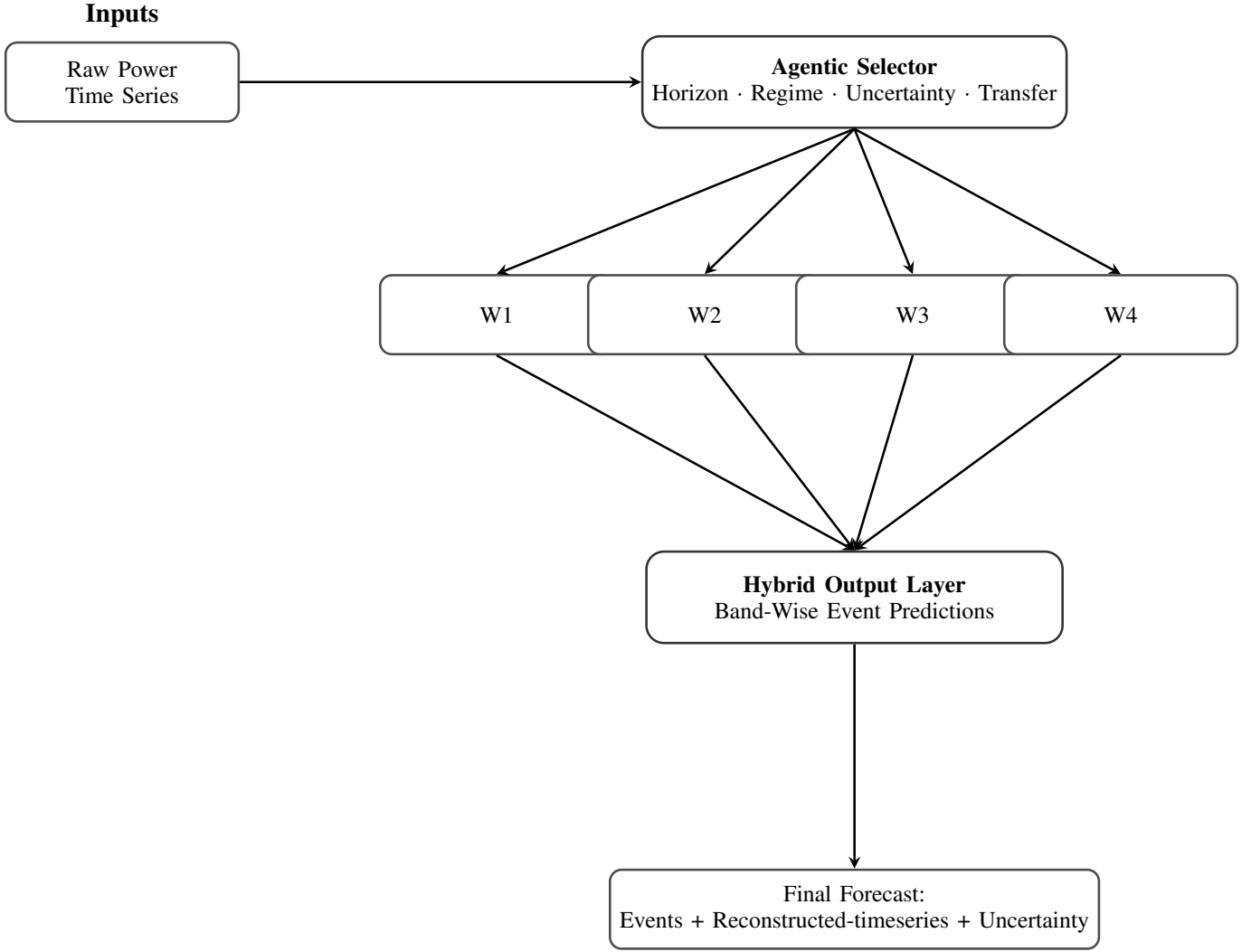
\begin{figure*}[htbp]
\centering
\resizebox{\textwidth}{!}{%
\begin{tikzpicture}[
    node distance=2.7cm,
    box/.style={
        rectangle,
        rounded corners=4pt,
        draw=black!70,
        fill=white,
        thick,
        minimum height=0.95cm,
        minimum width=2.8cm,
        font=\footnotesize,
        align=center
    },
    bigbox/.style={
        rectangle,
        rounded corners=6pt,
        draw=black!80,
        fill=white,
        thick,
        minimum height=1.1cm,
        minimum width=5.0cm,
        font=\footnotesize,
        align=center
    },
    arrow/.style={->, thick, >=stealth},
    title/.style={font=\bfseries\small}
]

%%%%%%%%%%%%%%%%%%%%%%%%%%%%%%%%%%%%%%%%%%%%%%%%%
% INPUT PIPELINE (LEFT)
%%%%%%%%%%%%%%%%%%%%%%%%%%%%%%%%%%%%%%%%%%%%%%%%%

\node[title] (inputTitle) at (-3.5,3.0) {Inputs};

\node[box] (raw) at (-3.5,2.2)
    {Raw Power\\Time Series};

%%%%%%%%%%%%%%%%%%%%%%%%%%%%%%%%%%%%%%%%%%%%%%%%%
% AGENTIC SELECTOR (STAGE A)
%%%%%%%%%%%%%%%%%%%%%%%%%%%%%%%%%%%%%%%%%%%%%%%%%

\node[bigbox] (selector) at (5.3,2.2)
{
    \textbf{Agentic Selector}\\
    Horizon · Regime · Uncertainty · Transfer
};

\draw[arrow] (raw.east) -- ++(0.8,0) |- (selector.west);

%%%%%%%%%%%%%%%%%%%%%%%%%%%%%%%%%%%%%%%%%%%%%%%%%
% WORKFLOWS (STAGE B)
%%%%%%%%%%%%%%%%%%%%%%%%%%%%%%%%%%%%%%%%%%%%%%%%%

\node[box] (W1) at (1.0,-0.6) {W1};
\node[box] (W2) at (3.5,-0.6) {W2};
\node[box] (W3) at (6.0,-0.6) {W3};
\node[box] (W4) at (8.5,-0.6) {W4};

% Straight arrows from selector to workflows
\draw[arrow] (selector.south) -- (W1.north);
\draw[arrow] (selector.south) -- (W2.north);
\draw[arrow] (selector.south) -- (W3.north);
\draw[arrow] (selector.south) -- (W4.north);

%%%%%%%%%%%%%%%%%%%%%%%%%%%%%%%%%%%%%%%%%%%%%%%%%
% OUTPUT FUSION
%%%%%%%%%%%%%%%%%%%%%%%%%%%%%%%%%%%%%%%%%%%%%%%%%

\node[bigbox] (merge) at (5.3,-4.0)
{
    \textbf{Hybrid Output Layer}\\
    Band-Wise Event Predictions
};

% Straight arrows from workflows to merge
\draw[arrow] (W1.south) -- (merge.north);
\draw[arrow] (W2.south) -- (merge.north);
\draw[arrow] (W3.south) -- (merge.north);
\draw[arrow] (W4.south) -- (merge.north);

%%%%%%%%%%%%%%%%%%%%%%%%%%%%%%%%%%%%%%%%%%%%%%%%%
% FINAL OUTPUT
%%%%%%%%%%%%%%%%%%%%%%%%%%%%%%%%%%%%%%%%%%%%%%%%%

\node[box] (final) [below=of merge]
{
    Final Forecast:\\
    Events + Reconstructed-timeseries + Uncertainty
};

\draw[arrow] (merge) -- (final);

\end{tikzpicture}}

\caption{Conceptual overview of the agentic forecasting system developed where $\{W_x\}$ are designed workflows}
\end{figure*}
%==================================================

Ramp events are commonly defined as intervals during which wind power changes exceed a specified threshold within a fixed time window. Their primary descriptors include amplitude, duration, rate of change, direction, onset time etc \cite{rba_sambeet}. The RBA$_\theta$ framework formalized this concept through slope-angle analysis of power trajectories, enabling systematic extraction of event-level features such as duration, intensity, and balance. Their analysis revealed that only a small fraction of samples typically correspond to ramp conditions, yet these intervals account for majority of the total power variance, highlighting their operational relevance.

Subsequent work extended this framework by introducing adaptive thresholding, Bayesian tuning, and Monte Carlo evaluation to handle non-stationary environments and quantify uncertainty \cite{my_rba}. This enhanced RBA$_\theta$ formulation bridged deterministic event detection with statistical reasoning, enabling more robust identification of ramps under varying atmospheric regimes. Nevertheless, predicting ramp events remains considerably more challenging than detecting them. The nonlinear relationship between wind speed and power output, together with mechanical lag, turbulence, and sensor noise, amplifies small input errors into large output deviations, particularly during rapid transitions.

A broad range of approaches has been proposed to address ramp detection and forecasting. Early threshold-based segmentation methods offer transparency and low computational cost but require manual tuning and exhibit limited transferability across sites. More recent studies combine signal decomposition techniques with deep learning architectures. Hybrid models incorporating wavelet transforms, Convolution Neural Networks, LSTMs, attention mechanisms, or variational decomposition have demonstrated improvements in ramp detection and short-horizon forecasting \cite{Li_pred, zhang_multi, xu_improved}. Similarly, \cite{sharp} found that while LSTM-RNN models achieved high accuracy on non-ramp intervals, they frequently underperformed during abrupt ramps, often missing extreme high readings and overfitting on stable data. This reinforces that ramp prediction operates under a distinct stochastic regime, where model assumptions suited for average dynamics break down in rare, high-impact deviations. In parallel, meteorological ensemble models such as NOAA HRRR have improved physical consistency but continue to exhibit substantial timing and amplitude errors at local scales \cite{bianco}. Review studies consistently report persistent challenges related to data imbalance, threshold sensitivity, and inadequate uncertainty representation \cite{zhang_det, li_review}. Likewise, \cite{eltohamy} shows that standard classification models trained on heavily imbalanced wind-power data can achieve high aggregate accuracy while exhibiting weak sensitivity to rare, high-impact ramp events. Even with imbalance-mitigation strategies, performance gains on extreme events remain limited, indicating that stochastic volatility and class skew impose structural constraints on generalisation. As a result, models optimised for average behaviour tend to underperform precisely in operationally critical scenarios.

Despite this progress, existing approaches remain fragmented across detection, forecasting, and uncertainty modeling. Event extraction is often treated as a preprocessing step rather than a co-evolving component of the forecasting system. This separation limits the ability to exploit shared structure between ramp onset, evolution, and dissipation, and motivates a transition toward integrated, event-aware forecasting systems.

%\IEEEpubidadjcol

\subsection{Knowledge Gap and Motivation}
\label{subsec:kg_motivation}

Although wind ramp research has advanced substantially, several fundamental gaps persist in event-level forecasting. First, there is no unified formulation of the ramp prediction problem. Existing studies variously frame ramps as binary classification tasks, magnitude regression problems, or two-stage detect-then-predict pipelines \cite{han_wind_pred, zhang_multi}. These formulations rely on differing definitions, thresholds, and evaluation metrics, which complicates comparison across datasets and masks true generalization capability. Reported improvements are often dataset-specific and difficult to transfer across sites or regimes.

Second, ramp events occur infrequently, typically representing fewer than 5\% of samples \cite{bianco}. This severe imbalance biases learning toward smooth operating conditions and degrades performance during rare but critical extremes. Even advanced deep architectures tend to minimize average error while misrepresenting ramp timing and magnitude, achieving low RMSE yet poor event-level reliability. This limitation reflects a structural mismatch between trajectory-based optimization objectives and event-driven operational needs.

Third, evaluation practices remain largely trajectory-centric. Metrics such as RMSE, MAE, and CRPS quantify average accuracy but fail to capture event timing, duration overlap, or directional correctness. Event-centric measures, including onset-time error, interval overlap, and event-wise precision or recall, are seldom reported. Moreover, probabilistic calibration diagnostics are rarely extended to discrete events, leaving the reliability of ramp forecasts insufficiently assessed.

Fourth, most forecasting pipelines are structurally rigid. Single-architecture models are applied uniformly across sites and regimes despite strong evidence that wind dynamics vary with terrain, atmospheric stability, and seasonal effects \cite{lochmann, he_ramp}. Few systems diagnose signal characteristics such as variance, entropy, or spectral composition and adapt their modeling strategy accordingly. The absence of autonomous workflow selection limits robustness under domain shift and constrains scalability.

Fifth, uncertainty is typically treated as a secondary output derived from ensemble spread or residual analysis. While probabilistic forecasting of continuous trajectories has matured, uncertainty at the event level remains largely unexplored. Existing methods rarely quantify confidence in event occurrence, timing, magnitude, or duration, despite the central role of such information in risk-aware grid operation \cite{he_ramp, rawson_multi}. The gap between trajectory-level uncertainty and event-level reliability remains unresolved.

Finally, interpretability and automation remain underdeveloped. Many deep models obscure physically meaningful descriptors such as slope, duration, and symmetry that are central to the RBA$_\theta$ methodology. At the same time, practical deployment is hindered by manual threshold tuning, site-specific retraining, and ad hoc preprocessing. End-to-end systems capable of autonomous execution and interpretable, event-aware outputs remain scarce. The emerging paradigm of agentic systems which are capable of introspection, self-configuration, and dynamic pipeline assembly promises transformative efficiency. LLM-based market agents have already shown practical autonomy in decision-intensive electricity markets, where a bidding-behavior agent and a sentiment agent improved 5-minute day-ahead spike and price predictions \cite{lu_agentic}. Translating this approach to physical time-series prediction together with other gaps collectively motivate a shift toward an integrated, agentic, event-centric forecasting paradigm that unifies detection, prediction, uncertainty quantification, and adaptive workflow selection. Within this context, the present work addresses the research questions presented in table \ref{tab:research_questions}. Together, these questions articulate the need for forecasting systems that are adaptive, interpretable, and uncertainty-aware, aligning predictive objectives with the operational realities of renewable-dominated power systems.

\begin{table*}[!t]
\centering
\caption{Research questions guiding the design and evaluation of the proposed agentic,
event-centric forecasting framework}
\label{tab:research_questions}
\renewcommand{\arraystretch}{1.25}
\begin{tabular}{p{0.08\textwidth} p{0.88\textwidth}}
\hline
\textbf{RQ} & \textbf{Research Question} \\
\hline
RQ1 &
Can a modular, agentic forecasting system be designed to select and combine modeling
workflows based on the statistical and spectral characteristics of the input signal,
instead of relying on a single fixed architecture? \\

RQ2 &
How can ramp detection and forecasting be jointly optimized to predict event occurrence
together with duration, magnitude, and direction within a unified learning framework? \\

RQ3 &
How can uncertainty be modeled, quantified, and evaluated at the event level, including
confidence in the timing, magnitude, and duration of predicted ramp events? \\

RQ4 &
What benefits arise from transitioning from trajectory-first forecasting to an
event-first, frequency-aware modeling approach with respect to accuracy, robustness,
reconstruction fidelity, and computational efficiency? \\

RQ5 &
How robust is an agentic, event-centric forecasting system across sites, temporal
resolutions, and data regimes, and to what extent can event semantics support
generalization beyond wind power time series? \\
\hline
\end{tabular}
\end{table*}

\section{Methodologies}
\label{sec:methodologies}

This section presents the methodological framework developed in this work. The overall system is organized as a set of complementary forecasting workflows, each representing a distinct modeling philosophy, and coordinated by an agentic selector that adapts to data characteristics and operational regimes. To establish a reliable reference point, we first introduce a transparent, trajectory-first baseline that reflects prevailing industrial practice. Subsequent approaches progressively incorporate event-centric learning, frequency-aware modeling, and adaptive workflow selection.

\subsection{Approach~1: Trajectory-First Forecasting with Post-hoc Event Extraction}

The first methodological component establishes an interpretable and fully reproducible baseline that reflects prevailing operational practice in wind power forecasting. The workflow follows a two-stage paradigm in which the wind-power trajectory is first forecast at the signal level and ramp-related events are subsequently extracted from the predicted series. Specifically, a Seasonal AutoRegressive Integrated Moving Average model with exogenous variables (SARIMAX)~\cite{alharbi_csala} is used to generate short-term forecasts, after which ramp and stationary periods are identified using the enhanced RBA$_\theta$ framework.

This decoupled design provides a transparent control condition against which more advanced hybrid and agentic methods can be evaluated. Errors introduced during forecasting and errors arising from event extraction remain separable, enabling direct attribution of performance degradation to either the trajectory model or the event logic. Such diagnostic clarity is essential for the later agentic selector, which must determine whether failures stem from inadequate signal modeling or from event-level interpretation.

Formally, given a historical wind-power series $W(t)$ and exogenous meteorological covariates $X(t)$ observed over a training interval, the forecaster produces a predicted trajectory $\widehat{W}(t)$ over the test horizon. Event predictions are obtained by applying RBA$_\theta$ directly to the forecast,
\begin{equation}
\label{eq:method1_events}
E_{\mathrm{pred}} = \operatorname{RBA}_{\theta}\!\bigl(\widehat{W}(t)\bigr),
\end{equation}
while ground-truth events $E_{\mathrm{act}}$ are extracted from the observed series using the same event logic. This symmetry enables event-level evaluation using overlap-based precision, recall, F1-score, and intersection-over-union (IoU).

For computational efficiency, the SARIMAX configuration adopted in this work employs no-differencing order $(p,d,q)=(1,0,1)$ and turned-off seasonal order $(P,D,Q,m)=(0,0,0,0)$. The model can be written as
\begin{equation}
\label{eq:sarimax_model}
y_t = c + \phi_1 y_{t-1} + \theta_1 \varepsilon_{t-1} 
      + \beta^\top X_t + \varepsilon_t,
\end{equation}
where $y_t$ denotes the raw wind-power series, $\varepsilon_t$ is white noise, and $X_t$ represents the selected exogenous variables. We compensation the omission of seasonal order through weather feature engineering and RBA$_\theta$ event descriptors. This formulation captures linear temporal dependencies, daily periodicity, and meteorological forcing while preserving coefficient interpretability.

Event extraction is performed using slope-based ramp logic central to RBA$_\theta$. Scale-normalized power gradients are computed using a symmetric difference,
\begin{equation}
\label{eq:grad_norm}
\nabla P(t) = \frac{P(t+\Delta t)-P(t-\Delta t)}{2\,\Delta t\, P_{\mathrm{rated}}},
\end{equation}
where $P_{\mathrm{rated}}$ is the turbine rated power. A ramp event is declared when the gradient magnitude exceeds an adaptive threshold $\tau(t)$ for a minimum duration,
\begin{equation}
\label{eq:ramp_condition}
|\nabla P(t)| > \tau(t), \qquad \text{duration} > T_{\min}.
\end{equation}

To accommodate non-stationary operating conditions, adaptive thresholds are employed. A statistically motivated rule defines
\begin{equation}
\label{eq:tau_stat}
\tau_{\mathrm{stat}}(t) = \mu_w(t) + k_\sigma \sigma_w(t),
\end{equation}
where $\mu_w(t)$ and $\sigma_w(t)$ denote rolling statistics of the signal or its slope proxy.
Stationary intervals are identified using complementary constraints on rolling slope, variance, and range, which substantially reduce false positives in low-variability regimes. Each detected event is characterized by onset time, duration, magnitude, direction, and auxiliary descriptors such as slope variance and symmetry~\cite{rba_sambeet}.

This baseline reflects common industrial forecasting pipelines, where signal-level predictions are produced upstream and interpreted downstream in terms of operationally meaningful events. Its statistical transparency, auditability, and explicit uncertainty propagation make it a suitable benchmark for evaluating the benefits of event-centric and agentic forecasting strategies introduced in subsequent sections.

\subsection{Approach~2: Direct Event Prediction Guided by RBA$_\theta$}

While the trajectory-first baseline provides transparency and diagnostic clarity, it inherently depends on accurate signal-level forecasting and therefore propagates trajectory uncertainty directly into event estimates. To address this limitation, the second methodological component adopts an event-first perspective in which ramp events are predicted directly, without explicitly forecasting the full power trajectory. This approach leverages the semantic structure exposed by RBA$_\theta$ and treats events as the primary prediction target rather than as post-hoc interpretations of a continuous signal.

The key premise of this approach is that many operational decisions depend on the timing, type, and persistence of events rather than on pointwise power values. By shifting the modeling focus from dense trajectories to discrete event semantics, this method reduces sensitivity to short-term signal noise and aligns more closely with the objectives of event-aware grid operation. It also establishes a complementary workflow that can be selected by the agentic controller when trajectory forecasting is unreliable, for instance under high turbulence or weak seasonality.

An initial attempt to realize this paradigm employed Survival Analysis, motivated by the natural correspondence between ramp forecasting and time-to-event modeling. Using Accelerated Failure Time formulations, the time until the next significant or stationary event was modeled as a function of RBA$_\theta$-derived descriptors such as event duration, magnitude, slope angle, and local variability. Despite its conceptual appeal, this approach exhibited limited empirical performance. Practical difficulties included sensitivity to sparse and irregular event spacing, instability of parametric assumptions, and frequent fallback to non-parametric estimators. These results indicated that classical survival models were too restrictive to capture the nonlinear and highly variable dynamics of wind-power ramps.

This observation motivated a transition to a more flexible yet still interpretable learning framework based on Random Forest classifiers. The adopted architecture decomposes direct event prediction into two sequential classification tasks. The first stage determines whether a significant event is occurring at a given time step, while the second stage assigns the significant event category, distinguishing between up and down ramps. This separation allows the model to learn onset dynamics and event semantics independently, without imposing distributional assumptions on event timing.

Let $x_t \in \mathbb{R}^d$ denote the feature vector at time $t$, composed of RBA$_\theta$ structural descriptors, meteorological covariates, and temporal encodings. Event occurrence is modeled as a probabilistic mapping
\begin{equation}
\label{eq:event_detection}
\widehat{y}_t = f(x_t) = \Pr(y_t = 1 \mid x_t),
\end{equation}
where $y_t \in \{0,1\}$ indicates the presence of an event. Conditioned on $\widehat{y}_t = 1$ i.e, for significant event time-steps, the type is predicted via
\begin{equation}
\label{eq:event_type}
\widehat{p}^{\mathrm{sig}}_t = g(x_t) = \Pr(c_t = \text{significant} \mid y_t = 1, x_t),
\end{equation}
with $c_t \in \{\text{significant}, \text{stationary}\}$. Both mappings are implemented using Random Forest ensembles trained on labels obtained from ground-truth RBA$_\theta$ segmentation. The hyperparameter used for this approach is described in table \ref{tab:rba_rf_params}.

\begin{table}[t]
\centering
\caption{Key hyperparameters used in the RBA$_\theta$-guided Random Forest
event prediction workflow}
\label{tab:rba_rf_params}
\renewcommand{\arraystretch}{1.15}
\setlength{\tabcolsep}{4pt}
\begin{tabularx}{\columnwidth}{l l X}
\hline
\textbf{Parameter} & \textbf{Symbol} & \textbf{Description} \\
\hline
Significant-event threshold multiplier 
& $\tau_{\text{sig}}$ 
& Multiplier controlling sensitivity of RBA$_\theta$ for detecting significant ramp events. \\

Stationary-event threshold multiplier  
& $\tau_{\text{stat}}$ 
& Threshold factor governing detection of stationary regimes in RBA$_\theta$. \\

Minimum event duration     
& $d_{\min}$ 
& Minimum temporal length required for a detected event to be retained. \\

Event-detection probability threshold  
& $\tau_{\text{event}}$ 
& Probability cutoff converting pointwise RF scores into event occurrence labels. \\

Minimum inter-event gap    
& $gap_{\min}$ 
& Minimum temporal separation enforced between consecutive events. \\

Temporal matching tolerance 
& $\delta$ 
& Allowed deviation when matching predicted events to ground-truth intervals. \\

Number of trees (event detection) 
& $B_{1}$ 
& Number of trees in the Random Forest used for event detection. \\

Maximum tree depth (event detection)      
& $D_{1}$ 
& Maximum depth of trees in the detection-stage Random Forest. \\

Number of trees (event classification) 
& $B_{2}$ 
& Number of trees in the Random Forest used for event-type classification. \\
\hline
\end{tabularx}
\end{table}

The detection stage is optimized using binary cross-entropy,
\begin{equation}
\label{eq:rf_det_loss}
\mathcal{L}_{\mathrm{det}}
= -\frac{1}{N}\sum_{t}
\bigl[
y_t \log \widehat{y}_t
+
(1-y_t)\log (1-\widehat{y}_t)
\bigr],
\end{equation}
while the type classifier minimizes an analogous loss over significant event points only. In the Random Forest setting, both objectives are approximated implicitly through impurity reduction, with the Gini index
\begin{equation}
\label{eq:gini}
G(S) = 1 - \sum_k p_k^2
\end{equation}
used to guide split selection, where $p_k$ denotes the class proportion within node $S$. Feature importance is computed as the average impurity decrease across trees, preserving interpretability of RBA$_\theta$-derived descriptors.

Pointwise predictions are subsequently aggregated into contiguous event intervals by thresholding $\widehat{y}_t$ and enforcing temporal continuity. Each predicted event is represented by its onset time, termination time, and class label, and matched to ground truth using a fixed temporal tolerance on boundaries. Event-level Precision, Recall, and F1-score are computed based on these matched intervals.

Although this Random Forest formulation is not the final architecture of the proposed system, it plays a critical role within the overall framework. It empirically demonstrates that direct event prediction is both feasible and substantially more effective than classical time-to-event modeling. Moreover, it provides a robust, interpretable workflow that can be autonomously selected by the agentic controller when signal-level forecasting is unreliable. As such, this approach constitutes a necessary intermediate step toward the fully event-centric, uncertainty-aware hybrid models introduced in the subsequent methodologies.

\subsection{Approach~3: Event-Aware Predictive Modelling via RBA$_\theta$--LSTM}
\label{subsec:approach3}

While Approach~1 follows a trajectory-first paradigm and Approach~2 demonstrates that
direct event prediction is feasible without explicit signal forecasting, both remain
limited in their ability to model \emph{event morphology as a structured temporal object}.
In particular, pointwise classification does not capture the coupled evolution of
event onset, magnitude, duration, and uncertainty across time.
Approach~3 introduces a fully event-aware, sequence-to-sequence predictive framework
that transforms the enhanced RBA$_\theta$ representation from a retrospective detector
into a forward-looking model that anticipates future event semantics.
Rather than forecasting the full signal or predicting isolated labels, the proposed
model directly predicts the temporal evolution of event attributes and reconstructs
the underlying trajectory only as a derived quantity when required.

\subsubsection{Design rationale}
\label{subsec:approach3_rationale}

The final architecture is shaped by three empirical observations.
First, event occurrence and event severity are governed by distinct temporal dynamics
and cannot be learned reliably from a single fixed-scale representation.
Second, wind-power events exhibit strong temporal clustering, requiring explicit
modelling of event-to-event dependence.
Third, different frequency components of the signal encode fundamentally different
physical processes, and forcing a single predictor to represent all scales leads to
systematic underfitting.
These observations motivate a design that combines explicit multi-resolution signal decomposition, causal modelling of event excitation, and frequency-aware prediction heads, while retaining the interpretability
and physical grounding of RBA$_\theta$.

\subsubsection{Final event-aware architecture}
\label{subsec:approach3_final_arch}

The final workflow integrates five tightly-coupled modules: 
enhanced RBA$_\theta$ event semantics, DWT-based multi-resolution decomposition, stratified bandit feature selection, a Hawkes layer modelling event clustering, and an encoder LSTM with frequency-aware prediction heads. After prediction of events, those are converted back into a reconstructed time series using the reconstruction method discussed in \ref{subsec:approach3_reconstruction}. A schematic overview of the final pipeline is shown in Fig.~\ref{fig:approach3_pipeline}.

\paragraph{Multi-resolution targets and inputs}
Let $P(t)$ denote the normalised power signal. A level-4 DWT decomposes $P(t)$ into an approximation component and detail components namely \texttt{Details\_1 (D1), Details\_2 (D2), Details\_3 (D4), and Details\_4 (D4)}. Enhanced RBA$_\theta$ is applied on each decomposed bands to obtain event-wise targets (onset, duration, magnitude, type) and timestep-level feature channels (event state, time-since-last-event, band statistics). These channels are concatenated into a multivariate sequence which the LSTM consumes over a fixed window length $S$ to predict event attributes over one or more horizons $\mathcal{H}$.

\paragraph{Horizon-aware stratified feature selection}
The expanded multi-resolution representation yields a large and heterogeneous feature
space.
To retain interpretability while avoiding over-parameterisation, the model employs
a horizon-aware, stratified multi-armed bandit (MAB) feature-selection mechanism
based on Thompson sampling.
Features are grouped into six semantic categories
(RBA$_\theta$, DWT, weather, power, temporal, and nonlinear),
each assigned a guaranteed quota to preserve domain balance.

Let $X \in \mathbb{R}^{N \times F}$ denote the engineered feature matrix and
$Y \in \{0,1\}^{N \times |\mathcal{H}|}$ the horizon-specific event labels.
Each feature $f$ is associated with a Beta posterior
\begin{equation}
\label{eq:mab_beta}
\theta_f \sim \mathrm{Beta}(\alpha_f,\beta_f),
\end{equation}
which represents the probability that feature $f$ contributes positively to
event prediction.
At each bandit round and for each prediction horizon, candidate subsets are sampled
within each category according to~\eqref{eq:mab_beta} and evaluated using a lightweight
Random Forest trained on temporally consistent splits.

The reward is defined as a horizon-balanced F1 score, and posterior parameters are
updated as
\begin{equation}
\label{eq:mab_update}
(\alpha_f,\beta_f) \leftarrow
\begin{cases}
(\alpha_f+3,\;\beta_f), & \text{high reward},\\
(\alpha_f+1,\;\beta_f), & \text{moderate reward},\\
(\alpha_f,\;\beta_f+1), & \text{otherwise},
\end{cases}
\end{equation}
encouraging exploration while progressively favouring features with stable
predictive utility.
After $R_{\mathrm{mab}}$ rounds, features are ranked by expected reward
$\mathbb{E}[\theta_f]$ and the top candidates per category are retained.

This procedure yields a compact feature set (approximately 75 predictors) that is
horizon-sensitive, category-balanced, and dominated by physically meaningful
ramp-scale components, particularly the D3 wavelet band.

\paragraph{Hawkes intensity as an auxiliary causal prior}
To account for bursty event arrivals, the model includes a Hawkes intensity term that increases event likelihood after recent events. Using detected event times $\{t_i\}$, the intensity is
\begin{equation}
\label{eq:hawkes_intensity}
\lambda(t)=\mu + \sum_{t_i < t} \alpha e^{-\beta (t-t_i)},
\end{equation}
where $\mu$ is a baseline rate and $\alpha,\beta$ control excitation strength and decay. The resulting intensity features are appended to the LSTM inputs, improving calibration of event occurrence under clustered regimes.

\paragraph{LSTM encoder \& Frequency-aware prediction heads}
Hawkes causality features concatenated with other input features are passed through two-stack LSTM encoders that produces a final latent representation of the input sequence $z$. Rather than a single output head, the latent sequence embedding is mapped to band-conditioned heads. This avoids forcing one output map to represent slow trend structure and ramp-scale dynamics simultaneously. In practice, the model predicts global event tasks (occurrence, type, time-to-event) and band-specific attributes (magnitude and duration contributions per band), then fuses these predictions into a final event tuple and, optionally, a reconstructed time series.

\paragraph{Multi-task objective with band-aware weighting}
The event-aware model is trained using a joint multi-task objective that combines
classification and regression losses across global event attributes and
frequency-specific components.
This formulation explicitly prioritises ramp-relevant structure while remaining
robust to heavy-tailed wind-power dynamics.

Let $z$ denote the latent representation produced by the LSTM encoder.
Global event attributes are predicted as
\begin{align}
p_{\mathrm{occ}} &= \sigma(W_{\mathrm{occ}} z), \\
p_{\mathrm{type}} &= \mathrm{softmax}(W_{\mathrm{type}} z), \\
t_{\mathrm{pred}} &= \mathrm{ReLU}(W_{\mathrm{time}} z),
\label{eq:global_event_predictions}
\end{align}
corresponding to event occurrence, event type, and time-to-event, respectively.
For each DWT band $b \in \{\mathrm{approx}, D4, D3, D2, D1\}$,
band-conditioned heads predict magnitude and duration contributions:
\begin{align}
m_b &= W^{(b)}_{\mathrm{mag}} z, \\
d_b &= \mathrm{ReLU}\!\bigl(W^{(b)}_{\mathrm{dur}} z\bigr),
\label{eq:bandwise_predictions}
\end{align}
where the frequency-aware parameterisation prevents slow-varying,
large-amplitude components from dominating ramp-scale dynamics.
To reflect the differing physical importance of each band, fixed importance
weights are assigned as
\[
w_{\mathrm{approx}} = 1.5, \quad
w_{d4} = 4.0, \quad
w_{d3} = 2.5, \quad
\]
\[
w_{d2} = 0.5, \quad
w_{d1} = 0.5.
\]

Magnitude and duration are optimised using a Huber loss
$L_{\mathrm{Huber}}(\cdot)$ with $\delta = 1$, yielding
\begin{align}
L_{\mathrm{mag}} &=
\frac{\sum_b w_b\, L_{\mathrm{Huber}}(m_b, m_b^{\ast})}{\sum_b w_b},
\label{eq:magnitude_loss}
\\
L_{\mathrm{dur}} &=
\frac{\sum_b w_b\, L_{\mathrm{Huber}}(d_b, d_b^{\ast})}{\sum_b w_b}.
\label{eq:duration_loss}
\end{align}
where $m_b$, $d_b$ are predictions and $m_b^\ast$, $d_b^\ast$ are ground truths.
Huber loss is employed to ensure robustness to extreme ramps and abrupt duration
variations, which are common in wind-power signals.
The global classification and timing objectives are defined as
\begin{align}
L_{\mathrm{occ}} &= \mathrm{BCE}(p_{\mathrm{occ}}, y_{\mathrm{occ}}), \\
L_{\mathrm{type}} &= \mathrm{CE}(p_{\mathrm{type}}, y_{\mathrm{type}}), \\
L_{\mathrm{time}} &= L_{\mathrm{Huber}}(t_{\mathrm{pred}}, t^{\ast}),
\label{eq:global_losses}
\end{align}

The final training objective is a weighted sum of all components:
\begin{equation}
L_{\mathrm{total}} =
2.0\,L_{\mathrm{occ}}
+
1.2\,L_{\mathrm{type}}
+
1.0\,L_{\mathrm{time}}
+
2.0\,L_{\mathrm{mag}}
+
2.0\,L_{\mathrm{dur}}.
\label{eq:total_loss}
\end{equation}

This weighting emphasises accurate event detection and event-property estimation
while maintaining balanced contributions across frequency bands.

\subsubsection{Event-Guided Time-Series Reconstruction}
\label{subsec:approach3_reconstruction}

The final component of Approach~3 is a reconstruction module that converts predicted
event semantics back into a physically plausible wind-power trajectory.
Reconstruction is not treated as a post-processing convenience, but as a
structural validation of whether the model has learned meaningful event dynamics
rather than merely discriminative patterns. In particular, high event-detection
accuracy without coherent reconstruction would indicate that the model detects
changes without understanding their magnitude, duration, or temporal context.

The central challenge arises from the event-centric formulation itself.
Events are discrete objects, whereas the target signal is continuous.
Naive reconstruction from predicted events alone creates large gaps between
consecutive events, since the model has no explicit representation of inter-event
dynamics. This problem is exacerbated in the multi-resolution setting, where
event extraction across wavelet bands discards fine-grained coefficient
information if not handled carefully.

To address this, reconstruction is formulated as a \emph{hybrid event--trend fusion}
problem that combines predicted event attributes with a slowly varying baseline
estimated from the approximation band. Crucially, no future ground-truth
information is used at inference time.

\paragraph{Baseline decomposition}
Let the predicted outputs of the event-aware model consist of event-level predictions for significant and stationary events
(onset, duration, magnitude, type),
and band-conditioned magnitude and duration estimates for the detail bands
(primarily $D3$ and $D4$, which dominate ramp dynamics).
The approximation band is treated separately, as it represents the
low-frequency trend of the signal.

\paragraph{Approximation-band handling}
Rather than reconstructing the approximation component solely from predicted
events (which proved unstable), the method leverages the slow temporal evolution
of the approximation band.
At inference time, the model produces a predicted approximation trajectory
$\widehat{A}_4(t)$.
In parallel, a reference approximation $\widetilde{A}_4(t)$ is obtained by
propagating the most recent past approximation coefficients forward,
which is feasible because $A_4$ varies slowly relative to the prediction horizons.
A trend-consistency check is then performed.
Both $\widehat{A}_4(t)$ and $\widetilde{A}_4(t)$ are smoothed using a wide Gaussian
filter to extract their low-frequency trend components.
If the correlation between the predicted and propagated trends exceeds a fixed
threshold, the predicted approximation is retained.
Otherwise, the propagated approximation is used as the baseline.
This mechanism prevents the reconstruction from drifting when the model fails
to learn long-term trends, while still allowing learned trends to dominate when
they are reliable.

\begin{algorithm}[t]
\caption{Event-Aware Forecasting with RBA$_\theta$--LSTM}
\label{alg:approach3}
\begin{algorithmic}[1]

\Require
Time series $P(t)$, meteorological covariates $W(t)$,  
wavelet $\psi$, decomposition level $L$,  
prediction horizon $h$, sequence length $S$.

\Ensure
Predicted events $\widehat{\mathcal{E}}$ and reconstructed signal $\widehat{P}(t)$.

\Procedure{EventAwareForecast}{$P(t), W(t)$}

\State Decompose $P(t)$ using DWT: $(A_4, D_4, D_3, D_2, D_1) \gets \textsc{DWT}(P,\psi,L)$

\For{each band $b \in \{A_4,D_4,D_3,D_2,D_1\}$}
    \State $\mathcal{E}_b \gets \textsc{RBA}_{\theta}(b)$
\EndFor

\State $\mathcal{E} \gets \textsc{FuseEvents}(\{\mathcal{E}_b\})$

\State $F \gets \textsc{FeatureEngineering}(P,W,\mathcal{E})$

\State $F^\star \gets \textsc{BanditSelect}(F)$

\State $\lambda(t) \gets \textsc{HawkesIntensity}(\mathcal{E})$

\State $Z \gets \textsc{LSTMEncode}(F^\star,\lambda)$

\State $\widehat{\mathcal{E}} \gets \textsc{PredictEvents}(Z)$

\State $\widehat{P}(t) \gets \textsc{Reconstruct}(\widehat{\mathcal{E}}, A_4)$

\State \Return $\widehat{\mathcal{E}}, \widehat{P}(t)$

\EndProcedure
\end{algorithmic}
\end{algorithm}

\paragraph{Event-driven detail reconstruction}
High-frequency structure is reconstructed exclusively from predicted events.
Only the detail bands that carry ramp-relevant dynamics (typically $D3$ and $D4$)
are used.
For each predicted event, its magnitude and duration are mapped to a
time-localised contribution in the corresponding band.
Between consecutive events, values are interpolated smoothly to close temporal
gaps, ensuring continuity without introducing artificial oscillations.
Stationary events constrain slope and variance in their intervals, preventing
spurious ramps in low-activity regions.
Lower-energy noise-dominated bands ($D1$ and $D2$) are not reconstructed from
predictions; their contribution is suppressed to avoid amplifying uncertainty.

\paragraph{Boundary anchoring and smoothing}
Even with correct band fusion, small baseline offsets at the temporal boundaries
can propagate across the reconstructed signal.
To prevent this, the reconstruction is anchored at the sequence boundaries using
a short transition window.
Within this window, reconstructed values are smoothly blended with boundary
reference values to eliminate discontinuities while preserving interior
predictions.
This produces a temporally coherent signal without sharp entry or exit artifacts.

\paragraph{Final reconstruction}
The complete reconstructed signal is obtained by summing the selected
approximation baseline with the reconstructed detail-band components and applying
the inverse discrete wavelet transform.
The result is a continuous trajectory whose large-scale trend is stable,
whose ramp structure is shaped by predicted events, and whose inter-event regions
are filled in a physically consistent manner.
This reconstruction strategy closes the loop of the event-aware forecasting
pipeline.
It ensures that predicted events correspond to a realizable time series,
thereby validating that the model has learned event morphology and temporal
dynamics rather than relying on classification shortcuts.

\begin{figure}[htbp]
\centering
\begin{tikzpicture}[
    node distance=0.95cm,
    box/.style={
        rectangle,
        rounded corners=6pt,
        draw=black!70,
        fill=white,
        thick,
        minimum width=7.0cm,
        minimum height=0.95cm,
        font=\small,
        align=center
    },
    arrow/.style={->, thick, >=stealth}
]

%%%%%%%%%%%%%%%%%%%%%%%%%%%%%%%%%%%%%%%%%%%%%%%%%
% TITLE
%%%%%%%%%%%%%%%%%%%%%%%%%%%%%%%%%%%%%%%%%%%%%%%%%

\node[font=\bfseries\large] (title) at (0,5.2)
{RBA--LSTM Event Prediction Full Pipeline};

%%%%%%%%%%%%%%%%%%%%%%%%%%%%%%%%%%%%%%%%%%%%%%%%%
% PIPELINE BLOCKS
%%%%%%%%%%%%%%%%%%%%%%%%%%%%%%%%%%%%%%%%%%%%%%%%%

\node[box] (input) at (0,4.2)
{Input Signal (Wind Power)};

\node[box] (dwt) [below=of input]
{\textbf{[1] DWT Decomposition}\\
$\rightarrow$ 5 Frequency Bands};

\node[box] (event) [below=of dwt]
{\textbf{[2] Per-Band Event Extraction}\\
(enhanced RBA$_\theta$)};

\node[box] (features) [below=of event]
{\textbf{[3] Feature Engineering}\\
RBA + DWT + Weather + Temporal+..};

\node[box] (mab) [below=of features]
{\textbf{[4] MAB Feature Selection}\\
(Stratified Thompson Sampling)};

\node[box] (hawkes) [below=of mab]
{\textbf{[5] Hawkes Process Layer}\\
(Event Causality Modeling)};

\node[box] (lstm) [below=of hawkes]
{\textbf{[6] LSTM Encoder}\\
(Sequence Learning)};

\node[box] (heads) [below=of lstm]
{\textbf{[7] Per-Band Prediction Heads}};

\node[box] (idwt) [below=of heads]
{\textbf{[8] iDWT Reconstruction}\\
(Frequency $\rightarrow$ Time Domain)};

%%%%%%%%%%%%%%%%%%%%%%%%%%%%%%%%%%%%%%%%%%%%%%%%%
% ARROWS
%%%%%%%%%%%%%%%%%%%%%%%%%%%%%%%%%%%%%%%%%%%%%%%%%

\draw[arrow] (input) -- (dwt);
\draw[arrow] (dwt) -- (event);
\draw[arrow] (event) -- (features);
\draw[arrow] (features) -- (mab);
\draw[arrow] (mab) -- (hawkes);
\draw[arrow] (hawkes) -- (lstm);
\draw[arrow] (lstm) -- (heads);
\draw[arrow] (heads) -- (idwt);

\end{tikzpicture}
\caption{Full pipeline of Approach~3}
\label{fig:approach3_pipeline}
\end{figure}
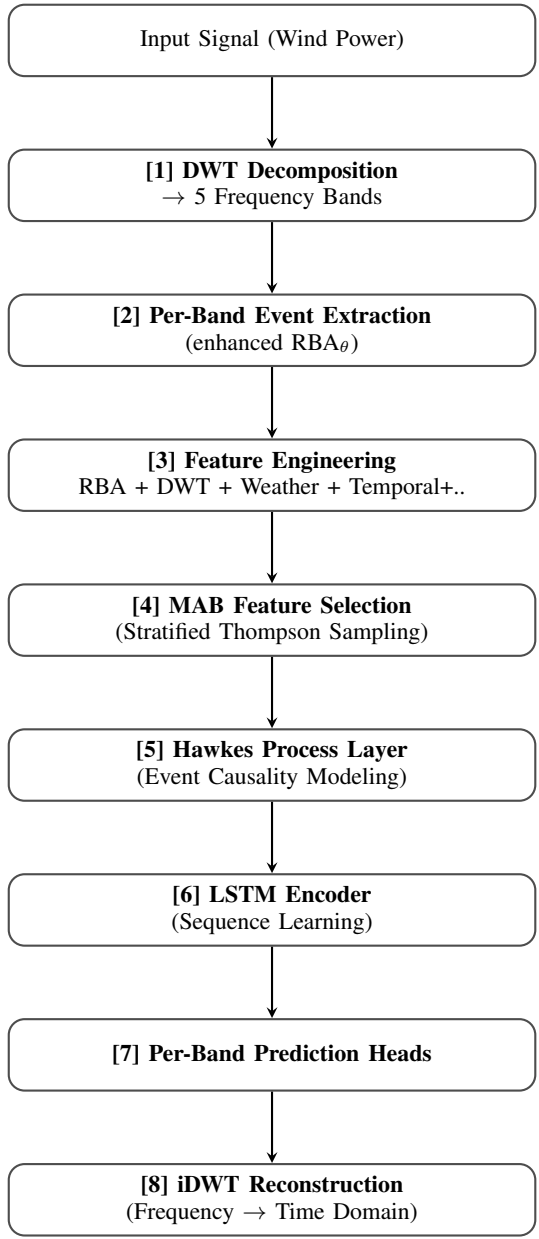

Algorithm~\ref{alg:approach3} provides a compact procedural specification of the
event-aware forecasting pipeline shown in Fig.~\ref{fig:approach3_pipeline},
formalizing the execution order and data dependencies without restating
architectural details.

\subsection{Approach~4: RBA$_\theta$ with Transformer-Based Direct Event Prediction}
\label{sec:approach4}

Approach~4 extends the event-first forecasting paradigm established in
Approach~3 by replacing the recurrent sequence encoder with a Transformer.
All upstream components like enhanced RBA$_\theta$ event extraction,
DWT-based multi-resolution processing, Hawkes-based temporal causality,
stratified MAB feature selection, band-aware prediction heads, and
event-guided reconstruction are \emph{identical} to Approach~3.
This strict inheritance ensures that any performance differences can be
attributed solely to the temporal modelling backbone.
The central question addressed by Approach~4 is therefore:
\emph{given the same event representation, features, horizons and objectives,
does a self-attention-based encoder provide complementary advantages over
recurrence for modelling wind-power event dynamics?}

\subsubsection{Transformer Encoder with Horizon-Specific Pathways}
\label{subsec:approach4_transformer}

Instead of an LSTM, Approach~4 employs a Transformer encoder to model
temporal dependencies across the event-aware feature sequence.
Given an input window
$\widetilde{X}_{1:S} \in \mathbb{R}^{S \times F}$,
the encoder computes
\begin{equation}
\begin{aligned}
Z^{(0)} &= \mathrm{PE}(\widetilde{X}), \\
Z^{(\ell)} &= \mathrm{EncoderBlock}\!\left(Z^{(\ell-1)}\right),
\qquad \ell = 1,\dots,L_{\mathrm{enc}} .
\end{aligned}
\end{equation}

where $\mathrm{PE}(\cdot)$ denotes learned positional encoding.
The final latent state $z_T$ is obtained via mean pooling over the temporal
dimension and is passed to the same frequency-aware prediction heads used
in Approach~3.

A key architectural addition is a \emph{dual-pathway design} for short horizons
($H \leq 1$\,h).
Short-horizon prediction is dominated by local persistence and derivative
patterns, whereas longer horizons depend on distributed precursors.
To prevent interference between these regimes, Approach~4 includes a
lightweight Transformer pathway with reduced depth and dimensionality
dedicated to short horizons, while the full encoder is used for longer horizons.
This separation stabilises optimisation and improves horizon-specific fidelity.

\subsubsection{Loss Function Adaptation}
\label{subsec:approach4_loss}

The multi-task structure and band-aware weighting scheme are inherited from
Approach~3.
However, the regression losses for magnitude and duration are modified to
better align with Transformer training stability.
For each band $B$ and horizon $H$, the loss components are defined as:
\begin{align}
L_{\mathrm{evt}}(B,H) &= \mathrm{BCE}
\bigl(\hat{y}_{\mathrm{evt}}^{(B)}(t+H),
      y_{\mathrm{evt}}^{(B)}(t+H)\bigr),\\
L_{\mathrm{mag}}(B,H) &=
\frac{1}{N}\sum_t
\bigl(
\hat{y}_{\mathrm{mag}}^{(B)}(t+H) -
y_{\mathrm{mag}}^{(B)}(t+H)
\bigr)^2,\\
L_{\mathrm{dur}}(B,H) &=
\frac{1}{N}\sum_t
\left|
\hat{y}_{\mathrm{dur}}^{(B)}(t+H) -
y_{\mathrm{dur}}^{(B)}(t+H)
\right|.
\end{align}

Mean squared error (MSE) is used for magnitude regression to preserve sensitivity
to amplitude deviations, while mean absolute error (MAE) is used for duration
to reduce bias toward over-smoothing.
The total loss per horizon is
\begin{equation}
\begin{aligned}
L_{\mathrm{total}}(H)
&=
\sum_{B}
\Bigl(
\lambda_1 L_{\mathrm{evt}}(B,H)
+ \lambda_2 L_{\mathrm{mag}}(B,H) \\
&\quad
+ \lambda_3 L_{\mathrm{dur}}(B,H)
\Bigr).
\end{aligned}
\end{equation}
with the same task and band weights as in Approach~3.
Approach~4 follows the same workflow as Approach~3 only the sequence encoder differs.

\begin{algorithm}[t]
\caption{Approach~4: Event-Aware Transformer Forecasting}
\label{alg:approach4}
\footnotesize
\begin{algorithmic}[1]

\Require 
Wind-power series $P(t)$, prediction horizons $\mathcal{H}$

\Ensure 
Event predictions (occurrence, type, timing, magnitude, duration)  
and optional reconstructed trajectory

\State Extract enhanced RBA$_\theta$ events and DWT bands
\State Construct Hawkes-based temporal features and statistical descriptors
\State Select $75$ features using horizon-aware stratified MAB

\For{each horizon $H \in \mathcal{H}$}
    \State Encode input sequence using Transformer encoder
    \State Predict event occurrence, type, timing
    \State Predict band-wise magnitude and duration
\EndFor

\State Optionally reconstruct trajectory via inverse DWT

\end{algorithmic}
\end{algorithm}

\begin{table}[htbp]
\caption{Key Hyperparameters for Approach~4}
\label{tab:approach4_hparams}
\centering
\footnotesize
\begin{tabular}{ll}
\toprule
Parameter & Value \\
\midrule
Transformer layers & 2 \\
Attention heads & 8 \\
Hidden dimension & 128 \\
Dropout rate & 0.45 \\
Sequence length & 48 hours \\
Feature dimension & 75 (MAB-selected) \\
Prediction horizons & \{1, 6, 12, 24\} h \\
Loss (mag / dur) & MSE / MAE \\
Optimiser & AdamW \\
\bottomrule
\end{tabular}
\end{table}

The shallow Transformer depth limits overfitting under limited data,
while high dropout counteracts attention overconfidence as shown in table \ref{tab:approach4_hparams}.
Sequence length and feature dimension mirror Approach~3 to preserve comparability.

\begin{table*}[t]
\caption{Comparison of the four approaches for wind-event prediction}
\label{tab:approach_comparison}
\footnotesize
\centering
\setlength{\tabcolsep}{3pt} % tighten spacing
\begin{tabularx}{\linewidth}{p{2.4cm} L L L L}
\toprule
\textbf{Aspect} &
\textbf{Approach 1} &
\textbf{Approach 2} &
\textbf{Approach 3} &
\textbf{Approach 4} \\
\midrule

\textbf{Core idea} &
Classical forecasting with post hoc event detection. &
Static event features with RF classification and survival-based timing. &
Deep multi-band sequence modelling with Hawkes-based temporal causality. &
Same as Approach~3, replacing LSTM with Transformer. \\

\addlinespace
\textbf{Frequency handling} &
None. &
Limited hand-crafted features. &
DWT-based multi-resolution processing. &
Same as Approach~3. \\

\addlinespace
\textbf{Causality modelling} &
None. &
Implicit (static RF). &
Hawkes intensities for temporal clustering. &
Self-attention and Hawkes causality. \\

\addlinespace
\textbf{Feature selection} &
Manual. &
Manual filtering. &
Stratified multi-armed bandit. &
Stratified multi-armed bandit. \\

\addlinespace
\textbf{Reconstruction} &
None. &
None. &
Event-guided reconstruction via DWT. &
Same as Approach~3. \\

\addlinespace
\textbf{Event prediction} &
Post hoc, threshold-based. &
Event-first with limited generalisation. &
Event-first with strong generalisation. &
Same as Approach~3. \\

\addlinespace
\textbf{Horizon support} &
Not implemented. &
Not implemented. &
Multi-horizon (1, 6, 12, 24 h). &
Multi-horizon (1, 6, 12, 24 h). \\

\addlinespace
\textbf{Magnitude / duration} &
Not explicitly modelled. &
Not modelled. &
Per-band magnitude and duration prediction. &
Same as Approach~3. \\

\addlinespace
\textbf{Zero-shot transfer} &
Not supported. &
Not supported. &
Moderate-to-good generalisation. &
Moderate-to-good generalisation. \\

\addlinespace
\textbf{Scalability} &
Low; per-site tuning required. &
Moderate; tree models remain site-specific. &
High. &
High. \\

\addlinespace
\textbf{Operational fit} &
High interpretability, low accuracy. &
Moderate interpretability and accuracy. &
High event fidelity with physical structure. &
Highly competitive operational relevance. \\

\bottomrule
\end{tabularx}
\end{table*}

\section{Agentic Workflow Orchestration for Wind Power Event Prediction}
\label{sec:agentic_orchestration}

The four workflows developed in Approaches 1-4 are intentionally complementary.
Each exhibits different strengths depending on dataset dynamics, horizon length,
and event regime. This motivates a final step beyond model design: a
self-learning orchestration layer that selects the most suitable workflow for a
given dataset and operational context.

Workflow selection was formulated as a \emph{contextual multi-armed bandit} (CMAB)
problem, where each workflow is an arm and the dataset itself provides the
context. Let
\[
\mathcal{W}=\{w_1,w_2,w_3,w_4\}
\]
denote the candidate workflows, corresponding to Approach 1-4 respectively. For any dataset $D$, compact statistical and dynamical fingerprints were extracted capturing volatility, stationarity, trend strength, regime dynamics, entropy,
multi-lag autocorrelation and predictability indicators. This context enables
knowledge transfer across heterogeneous wind farms without site-specific
hand-tuning.

\subsection{Dataset Similarity and Experience Retrieval}
\label{subsec:agentic_similarity}

To reuse prior executions, similarity between two datasets $D_1$ and
$D_2$ using cosine similarity in the context space were quantified:
\begin{equation}
\label{eq:cosine_similarity}
\mathrm{sim}(D_1,D_2)
=
\frac{\mathbf{C}(D_1)^\top \mathbf{C}(D_2)}
{\|\mathbf{C}(D_1)\|_2\,\|\mathbf{C}(D_2)\|_2}.
\end{equation}
Given a similarity threshold $\tau\in(0,1)$, a neighbourhood of past
executions from a persistent experience base $\mathcal{E}$ were retrieved:
\begin{equation}
\label{eq:experience_retrieval}
\mathcal{E}_{\mathrm{sim}}
=
\left\{
E_j\in\mathcal{E} \;:\;
\mathrm{sim}\!\left(D_j, D\right)>\tau
\right\}.
\end{equation}
Each record $E_j$ stores the dataset context, the selected workflow, evaluation
metrics, timestamp and a scalar reward $R_j$ (defined below). This provides a
lightweight mechanism for context-conditioned decision making.

\subsection{Contextual Reward and Utility}
\label{subsec:agentic_reward}

A composite reward that jointly reflects forecasting fidelity,
event-level detection quality and computational efficiency was defined. For an execution of
workflow $w_i$ on dataset $D$, the reward is
\begin{equation}
\label{eq:reward_def}
\begin{aligned}
R(w_i,D)
&=
0.4\,R^2_{\mathrm{test}}
+
0.2\left(1-\frac{\mathrm{MAE}_{\mathrm{test}}}{500}\right) \\
&\quad
+
0.3\,F1_{\mathrm{events}}
+
0.1\left(1-\frac{t_{\mathrm{exec}}}{3600}\right).
\end{aligned}
\end{equation}

where $R^2_{\mathrm{test}}$ and $\mathrm{MAE}_{\mathrm{test}}$ measure predictive
accuracy, $F1_{\mathrm{events}}$ measures event detection quality, and
$t_{\mathrm{exec}}$ is total execution time in seconds. The constants normalise
each component to comparable scale and emphasise event-relevant performance.

\subsection{Weighted Value Estimation from Similar Executions}
\label{subsec:agentic_q}

For each workflow $w_i$, a context-conditioned expected utility by
a weighted average over similar executions was estimated:
\begin{equation}
\label{eq:q_estimate}
Q(w_i,\mathbf{C})
=
\frac{\sum\limits_{E_j\in\mathcal{E}_{\mathrm{sim}}(w_i)} \gamma_j\,R_j}
{\sum\limits_{E_j\in\mathcal{E}_{\mathrm{sim}}(w_i)} \gamma_j},
\end{equation}
where $\mathcal{E}_{\mathrm{sim}}(w_i)$ denotes similar executions using workflow
$w_i$, and $\gamma_j$ is a reliability weight that down-weights stale,
synthetic or low-confidence evidence.
$\gamma_j$ is a product of three interpretable factors:
\begin{equation}
\label{eq:gamma_factorization}
\gamma_j
=
\gamma^{(\mathrm{rec})}_j
\;\gamma^{(\mathrm{src})}_j
\;\gamma^{(\mathrm{conf})}_j.
\end{equation}
The recency term applies exponential decay so that recent evidence dominates:
\begin{equation}
\label{eq:gamma_recency}
\gamma^{(\mathrm{rec})}_j
=
\exp\!\left(-\frac{\Delta t_j}{30}\right),
\end{equation}
where $\Delta t_j$ is the age of execution $E_j$ measured in days. The source
term ensures that synthetic bootstrap knowledge decays as real data accumulates:
\begin{equation}
\label{eq:gamma_source}
\gamma^{(\mathrm{src})}_j
=
\alpha^{n_{\mathrm{real}}},
\qquad
\alpha=0.95,
\end{equation}
where $n_{\mathrm{real}}$ is the number of real executions stored so far.
Finally, the confidence term penalises workflows with limited evidence in the
retrieved neighbourhood:
\begin{equation}
\label{eq:gamma_conf}
\gamma^{(\mathrm{conf})}_j
=
\min\!\left(\frac{n_i}{5},\,1\right),
\end{equation}
where $n_i$ is the number of similar executions available for workflow $w_i$.
Together, \eqref{eq:gamma_factorization}--\eqref{eq:gamma_conf} yield a
practical estimator that is robust to stale logs, cold-start priors and sparse
coverage.

\subsection{Exploration--Exploitation with UCB and Adaptive $\epsilon$}
\label{subsec:agentic_exploration}

Given $Q(w_i,\mathbf{C})$, the ideal choice is
\begin{equation}
\label{eq:argmax_expected_reward}
w^\star
=
\arg\max_{w_i\in\mathcal{W}}
\mathbb{E}\!\left[R(w_i,D)\mid \mathbf{C}(D)\right],
\end{equation}
which was approximated by maximising $Q(w_i,\mathbf{C}(D))$ when exploiting.
To avoid over-commitment under uncertainty, the agent also explores using an
upper-confidence strategy:
\begin{equation}
\label{eq:ucb}
\mathrm{UCB}(w_i,\mathbf{C})
=
Q(w_i,\mathbf{C})
+
c\sqrt{\frac{\ln N}{n_i}},
\qquad
c=2.0,
\end{equation}
where $N=\sum_i n_i$ is the total number of similar executions and $n_i$ is the
count for workflow $w_i$. This encourages exploration when $n_i$ is small.
Exploration is controlled using an adaptive $\epsilon$-greedy policy:
\begin{equation}
\label{eq:epsilon_adaptive}
\epsilon(t)
=
\epsilon_{\mathrm{base}}
\,f_{\mathrm{contr}}
\,f_{\mathrm{local}}
\,f_{\mathrm{cons}}
\,f_{\mathrm{data}},
\qquad
\epsilon_{\mathrm{base}}=0.2,
\end{equation}
with $\epsilon(t)\in[0.05,0.60]$. The multiplicative factors increase
exploration when contradictions are detected between synthetic and real
evidence, when selection concentrates on a single workflow, or when the agent
has repeatedly exploited without testing alternatives. This improves
robustness under dataset shift and prevents premature convergence.

\subsection{Contradiction Detection and Bootstrap Decay}
\label{subsec:agentic_contradiction}

To prevent bootstrap bias, contradictions between synthetic bootstrap executions and real executions are explicitly detected. Let $\mathcal{S}$ and $\mathcal{R}$
denote the subsets of retrieved similar executions originating from synthetic
priors and real runs, respectively; then the source-specific best workflow is defined as:
\begin{equation}
\label{eq:best_synthetic_real}
w^\star_{\mathcal{S}}
=
\arg\max_{w_i\in\mathcal{W}} \mu_{\mathcal{S}}(w_i),
\qquad
w^\star_{\mathcal{R}}
=
\arg\max_{w_i\in\mathcal{W}} \mu_{\mathcal{R}}(w_i),
\end{equation}
where $\mu_{\mathcal{S}}(w_i)$ and $\mu_{\mathcal{R}}(w_i)$ denote mean rewards
under each source. A contradiction is flagged when
\[
w^\star_{\mathcal{S}}\neq w^\star_{\mathcal{R}},
\quad
\left|\mu_{\mathcal{S}}(w^\star_{\mathcal{S}})-\mu_{\mathcal{R}}(w^\star_{\mathcal{R}})\right|>0.10,
\quad
|\mathcal{R}|\ge 3.
\]
When this occurs, Synthetic evidence is discarded for the current decision and
increase exploration, ensuring that real evidence dominates selection as soon
as it is sufficiently informative.

\subsection{Learning Loop and Persistent Memory}
\label{subsec:agentic_learning_loop}

After selecting a workflow and executing it, the agent computes
$R(w^\star,D)$ using \eqref{eq:reward_def} and appends a new record to the
experience base $\mathcal{E}$. This yields a self-improving loop: each run
refines future estimates via \eqref{eq:q_estimate}, while the bootstrap influence
decays automatically through \eqref{eq:gamma_source}. In effect, the system
does not only learn event predictors; it learns \emph{which} predictor to trust
under which dataset conditions. This closes the methodological loop of the paper
by integrating Approaches~1--4 into a unified, adaptive decision-making layer
that targets robust wind-power event prediction under heterogeneous operating
regimes.

\section{Results and Analysis}
\label{sec:results_analysis}

\subsection{Dataset Description and Environmental Setup}
\label{subsec:data_and_setup}

The proposed event-centric and frequency-aware forecasting framework is evaluated
using multiple offshore wind datasets that differ in geographical location,
meteorological regimes, and operational characteristics. The usage of the datasets are summarized in Table \ref{tab:dataset_roles}.
This design supports both in-distribution evaluation and strict zero-shot
generalisation to unseen wind farms.

The primary training and validation dataset corresponds to the Baltic Eagle
offshore wind farm and is obtained from the publicly available
``40 Years of European Offshore Wind'' dataset generated using the
Copernicus Climate Change Service (C3S) ERA5 reanalysis
\cite{grothe}.
The dataset provides 40 years (1980--2019) of hourly observations including
hub-height wind information, wind direction, surface roughness, and turbine- and
farm-level power output.
Model development, feature selection, and hyperparameter tuning are performed
exclusively on Baltic Eagle using a chronological split:
70\% training, 15\% validation, and 15\% test, ensuring no temporal leakage.
To evaluate zero-shot transferability, two additional offshore wind farms from
the same repository are used without any retraining or adaptation.

One of them is Baie de Saint-Brieuc (English Channel) data that exhibits strong mesoscale variability
influenced by Atlantic inflow and boundary-layer transitions, while the other one i.e, London Array
(Thames Estuary) dataset reflects frequent frontal activity and complex turbulence and
wake interactions.
For both transfer datasets, the evaluation strictly uses only variables provided
in the original dataset to preserve the integrity of the zero-shot setting.
An additional Estonian coastal capacity-factor dataset with eight-turbines, also used in earlier RBA$_\theta$
studies \cite{my_rba, rba_sambeet}, is employed only during initial baseline
development and algorithm validation and is excluded from all final learning-based
evaluations due to its synthetic nature and limited temporal depth.

Power output is normalised by dividing by the nominal (rated) capacity to obtain
a unit-scale series consistent across sites.
All experiments are conducted on a standard desktop system equipped with a
quad-core Intel CPU operating at 2.0\,GHz and 16\,GB of RAM.
The implementation is written in Python~3.9. The complete implementation is available at \href{https://github.com/sambeets/rbaTheta}{this Github link}.

\begin{table}[!t]
\caption{Summary of dataset roles across the workflow}
\label{tab:dataset_roles}
\centering
\footnotesize
\begin{tabularx}{\linewidth}{lXXX}
\toprule
Dataset & Model Training & Event Extraction & Zero-Shot Transfer \\
\midrule
Baltic Eagle & Yes & Yes & -- \\
Baie de Saint-Brieuc & -- & Yes & Yes \\
London Array & -- & Yes & Yes \\
Eight-Turbine Dataset & Initial development & Baseline only & -- \\
\bottomrule
\end{tabularx}
\end{table}

\subsection{Analysis and Discussions}
\label{subsec:results}

This section evaluates the proposed event-aware, frequency-informed forecasting framework
introduced in Section~\ref{sec:methodologies}.  
The analysis mostly focuses on the two competitive deep workflows (Approach~3 and Approach~4),
while classical baselines are used to contextualise performance gains. 

\begin{figure*}[!t]
    \centering
    \includegraphics[width=\textwidth]{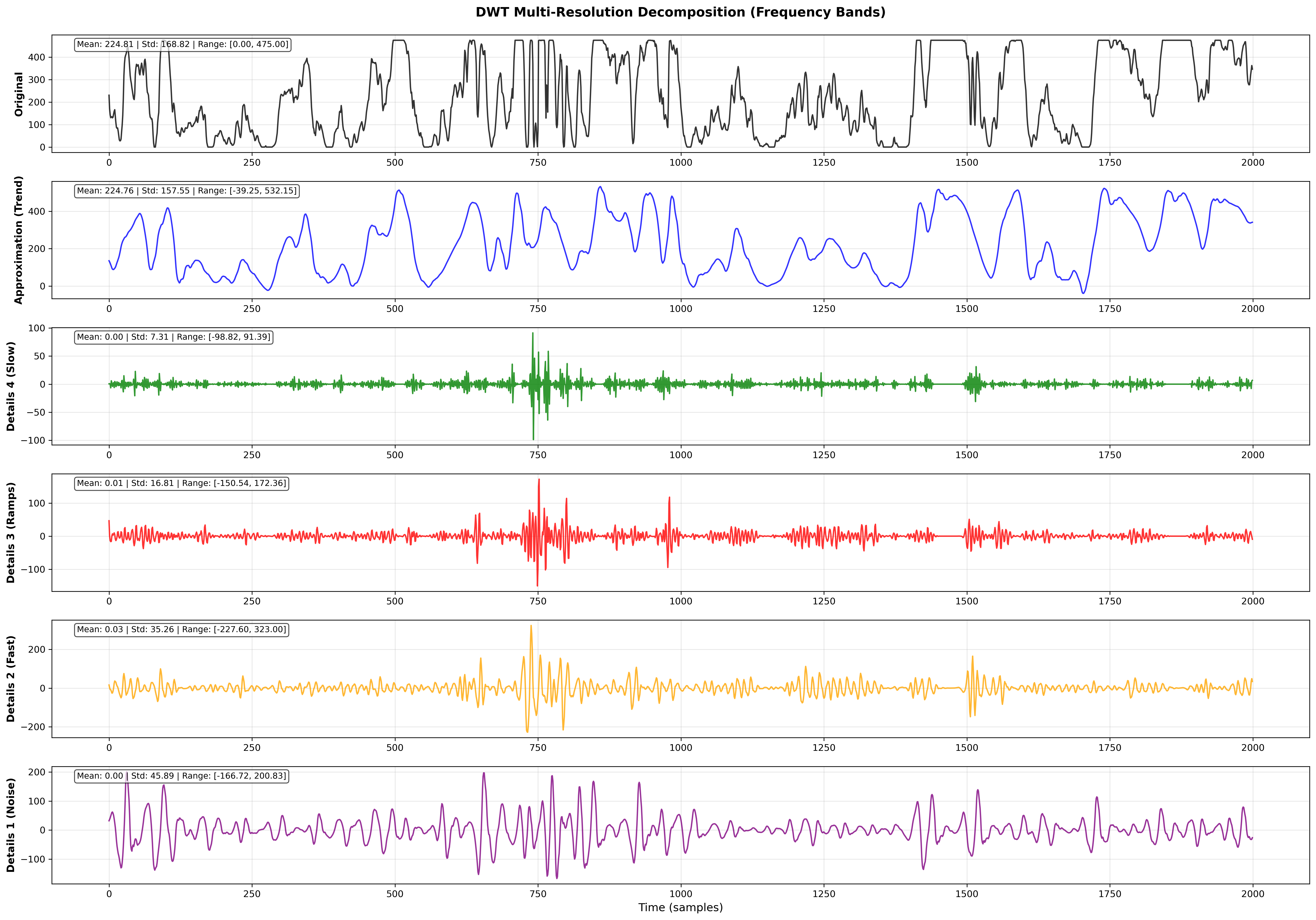}
    \caption{DWT decomposition of the wind-power time series into approximation and detail bands.}
    \label{fig:dwt-decomposition}
\end{figure*}

To expose the temporal structure of wind-power variability across scales, the power
time series is decomposed using Discrete Wavelet Transform (DWT) into one
approximation component and four detail components, as illustrated in
Fig.~\ref{fig:dwt-decomposition}.
The original series exhibits a mean of $224.81$~MW, a standard deviation of
$168.82$~MW, and a range spanning $[0, 475]$~MW, indicating substantial variability
driven by atmospheric dynamics.

The approximation band captures slow-varying background behaviour, with statistics
(mean $224.76$~MW, std $157.55$~MW) closely matching the original signal.
Its wide amplitude range reflects long-term regime shifts and seasonal structure,
but it carries limited information about rapid ramp events.
In contrast, the detail bands isolate progressively faster dynamics.
Among them, the $D3$ band (approximately 2--8~h) exhibits the largest variance
(std $16.81$~MW) and the widest range among the mid-frequency components,
indicating that it concentrates most ramp-related energy.
The $D4$ band (8--16~h) contributes smoother sub-diurnal fluctuations,
while the higher-frequency bands ($D2$, $D1$) show increasing variance dominated
by short-term oscillations and noise rather than coherent ramp structure.
These statistics justify focusing predictive attention on mid-frequency bands for
event characterisation, while treating the highest-frequency components primarily
as noise carriers and the approximation band as a reconstruction scaffold.

Across all experimental workflows, adaptive feature selection consistently highlights
a common set of physically meaningful predictors.
Event-derived quantities (such as event magnitude, duration, density and timing)
form the most stable contributors to event occurrence and classification.
Wavelet-based features from the mid-frequency bands, particularly variance, energy
and cross-band ratios, repeatedly emerge as strong indicators of ramp intensity and
persistence.
Meteorological drivers related to pressure gradients, wind-speed evolution and
synoptic forcing provide complementary context, while temporal encodings
(hour-of-day, cyclic seasonality) support alignment with diurnal and seasonal patterns.
Nonlinear interaction terms further capture compound effects that are not apparent
from marginal statistics alone.
These observations indicate that reliable wind-event prediction
relies on a balanced combination of event semantics, mid-scale frequency structure
and physically grounded atmospheric cues.
Adaptive selection mechanisms, such as bandit-based exploration, consistently converge
towards this combination, reinforcing that the identified features reflect genuine
wind-power physics rather than artefacts of a particular model or dataset.

Based on this foundation, forecasting results are mostly reported along six axes: event prediction accuracy, event magnitude and duration estimation, signal reconstruction fidelity, computational costs, ablation studies, and zero-shot transferability across wind farms.

\subsubsection{Event Prediction}
\label{subsubsec:event-pred}
Event prediction quality is assessed mostly using precision, recall, and F1-score across forecasting horizons from 1\,h to 24\,h.
Precision is emphasised due to its operational importance in ramp-alert systems.

Approach~1 provides a strong interpretable reference,
achieving $\mathrm{F1}$ of 0.91 on the Baltic Eagle dataset with low false-alarm rates.
This establishes an upper-bound for rule-based post-hoc extraction
and motivates direct event prediction.
Approach~2 formulates event prediction as a direct classification problem
using a two-stage RF--RBA$_\theta$ pipeline.
The survival-analysis variant exhibits a clear performance ceiling with
the best Kaplan-Meier fallback model configuration attaining a
$\mathrm{F1}$ score of 0.41 with precision of $\approx0.38$ and the recall was $\approx0.46$. Weibull-AFT completely failed to run, mainly due to mathematical instability (failed matrix inversion), high collinearity among covariates, and severe event sparsity.

\begin{figure*}[!t]
\centering
\includegraphics[
    width=\textwidth,
    keepaspectratio
]{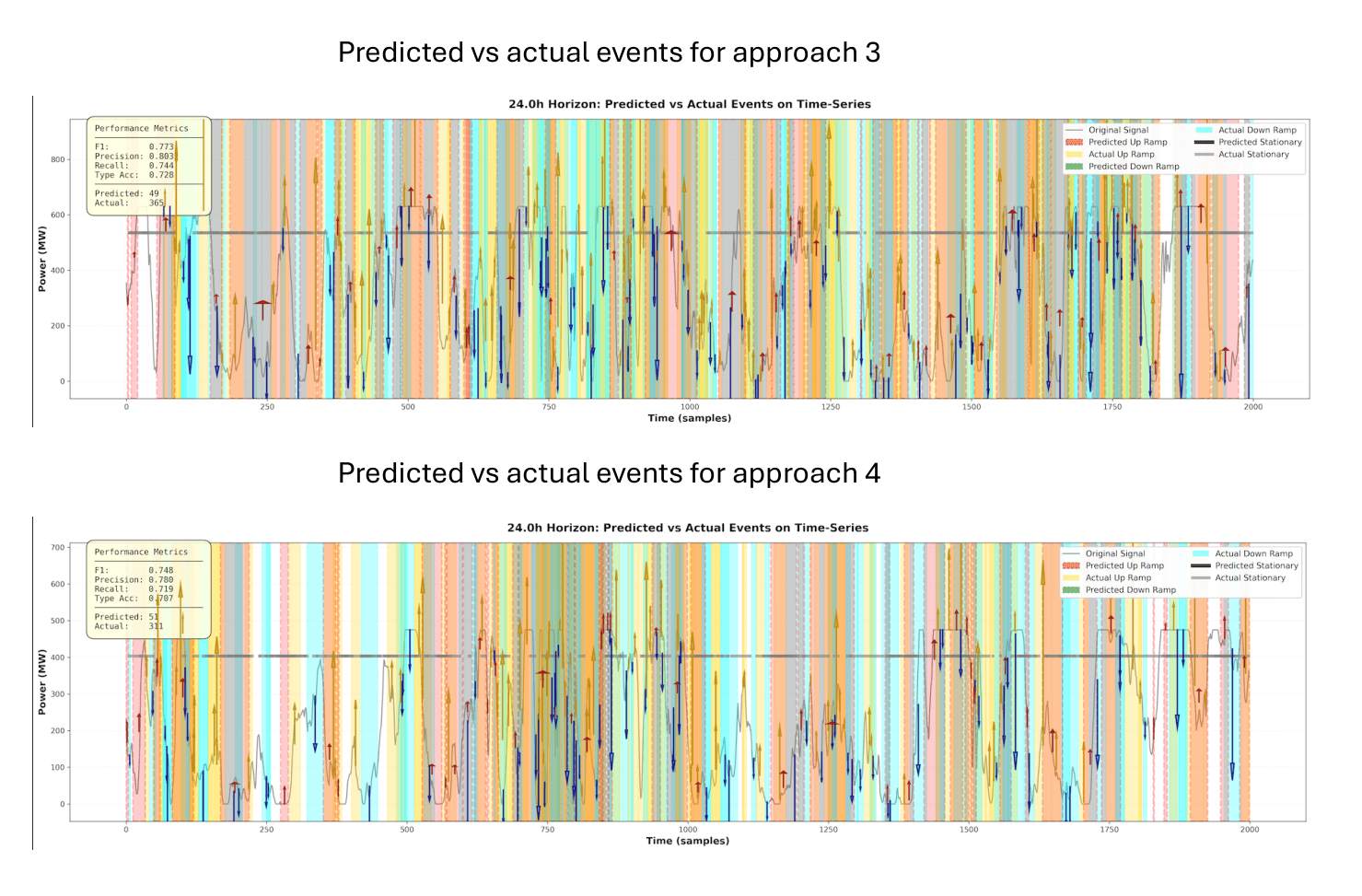}
\caption{Comparison of predicted and true events for 24-hour horizon using Approach~3 \& 4}
\label{fig:pred-events-app3}
\end{figure*}

Replacing survival modelling with a pointwise Random Forest classifier
substantially improves predictive capacity. On the Baltic Eagle test split, the RF baseline achieves a precision of 0.76 with $83.5\%$ relative improvement in $\mathrm{F1}$-score over the survival baseline.
This establishes Approach~2 as the strongest non-deep reference model
for direct event occurrence and type prediction.
A class-wise analysis further reveals asymmetric behaviour.
For significant ramps, the classifier is conservative,
achieving precision $\approx0.85$ and recall $\approx0.62$, while for stationary periods it prioritises coverage,
with recall $\approx0.93$ and precision $\approx0.70$.
This precision–recall trade-off aligns with operational priorities
and provides a meaningful quantitative baseline against which
the deep event-predictive workflows (Approaches~3 and~4) are evaluated.
For Approach~3, event prediction performance improves monotonically with horizon.
At short horizons (1\,h), recall is high but precision is limited due to noise sensitivity.
At longer horizons (12-24\,h), the model achieves balanced performance
with relatively high precision of $\approx0.84$ in context of direct event prediction, indicating reliable early identification of coherent ramp structures. Approach~4 exhibits a more conservative prediction profile. While short-horizon performance is weaker,
precision improves steadily with horizon and peaks at $\approx0.78$ at 24\,h.
Compared to the LSTM, as shown in Fig~\ref{fig:pred-events-app3}, the Transformer produces smoother and less fragmented event sequences,
at the cost of missing some shorter ramps. This qualitative comparison confirms the quantitative trends:
short-horizon predictions are reactive,
while longer horizons yield stable and operationally useful segmentation.

% ======================================================
\subsubsection{Magnitude and Duration Prediction}
\label{subsubsec:mag_dur_results}

\begin{figure}[htbp]
\centering
\includegraphics[width=\linewidth]{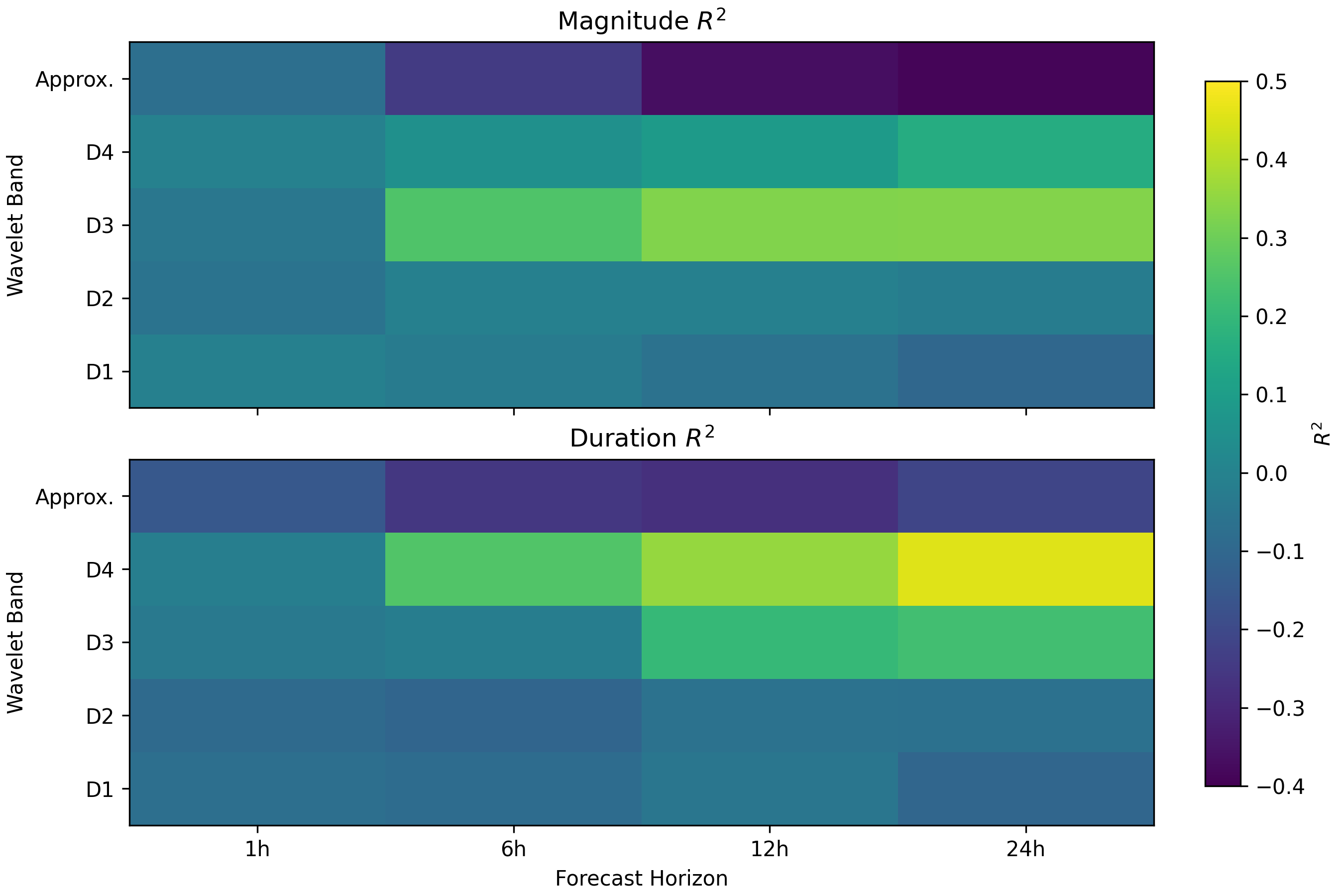}
\vspace{0.5em}
\includegraphics[width=\linewidth]{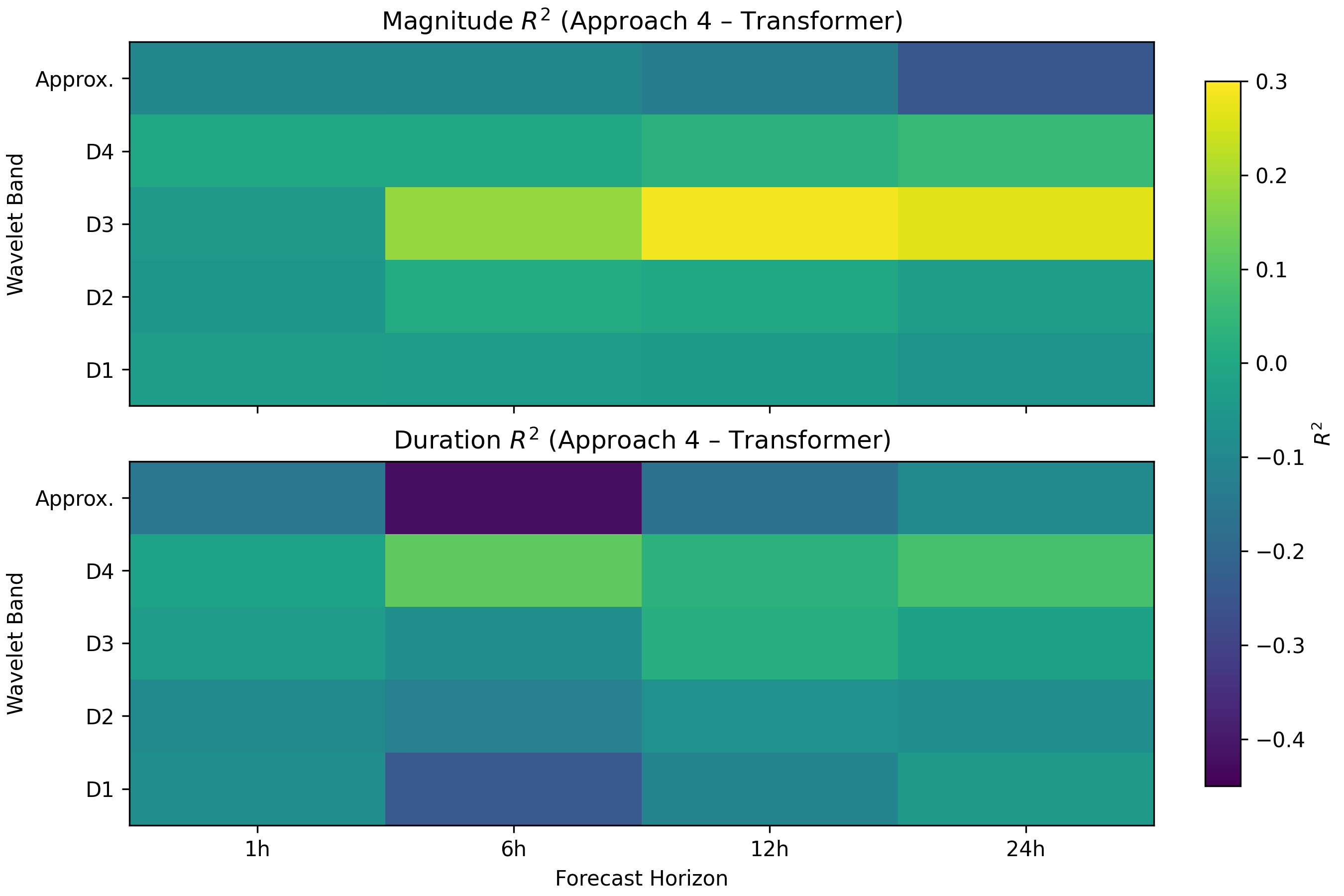}
\caption{Per-band $R^2$ heatmaps for magnitude (top) and duration (bottom).
Approach~3 shown above, and Approach~4 below}
\label{fig:perband-r2-comparison}
\end{figure}

Event severity is quantified through magnitude and duration regression.
Performance is evaluated using the coefficient of determination ($R^2$)
computed separately for each DWT frequency band as shown in Figure~\ref{fig:perband-r2-comparison}.

In case of Approach~3, Mmgnitude prediction exhibits strong frequency selectivity.
The mid-frequency ramp band (\texttt{Details\_3}) dominates performance, as it contains the most ramp energies,
reaching $R^2$ of $0.33$ at 24\,h.
High-frequency bands remain non-predictive, indicating effective noise suppression.
Duration prediction is more challenging but improves with horizon,
with \texttt{Details\_4} reaching $R^2\approx0.46$ at 24\,h. On the other hand, approach~4 preserves the same frequency structure but with smoother behaviour.
Magnitude prediction in \texttt{Details\_3} reaches $R^2\approx0.27$ at long horizons,
slightly below the LSTM peak but more stable across horizons.
Duration prediction remains weaker overall, suggesting that recurrent inductive bias better captures persistence. These results demonstrate that

(i) ramp magnitude is governed primarily by mid-frequency dynamics,

(ii) duration requires fine-scale temporal continuity, and

(iii) both models correctly suppress low-information bands.

% ======================================================
\subsubsection{Signal Reconstruction Performance}
\label{subsubsec:reconstruction_results}

Reconstruction evaluates whether predicted events encode sufficient information to recover physically meaningful power trajectories. It has been performed using the method discussed in Subsection~\ref{subsec:approach3_reconstruction}.

\begin{figure*}[!t]
\centering
\includegraphics[width=\textwidth]{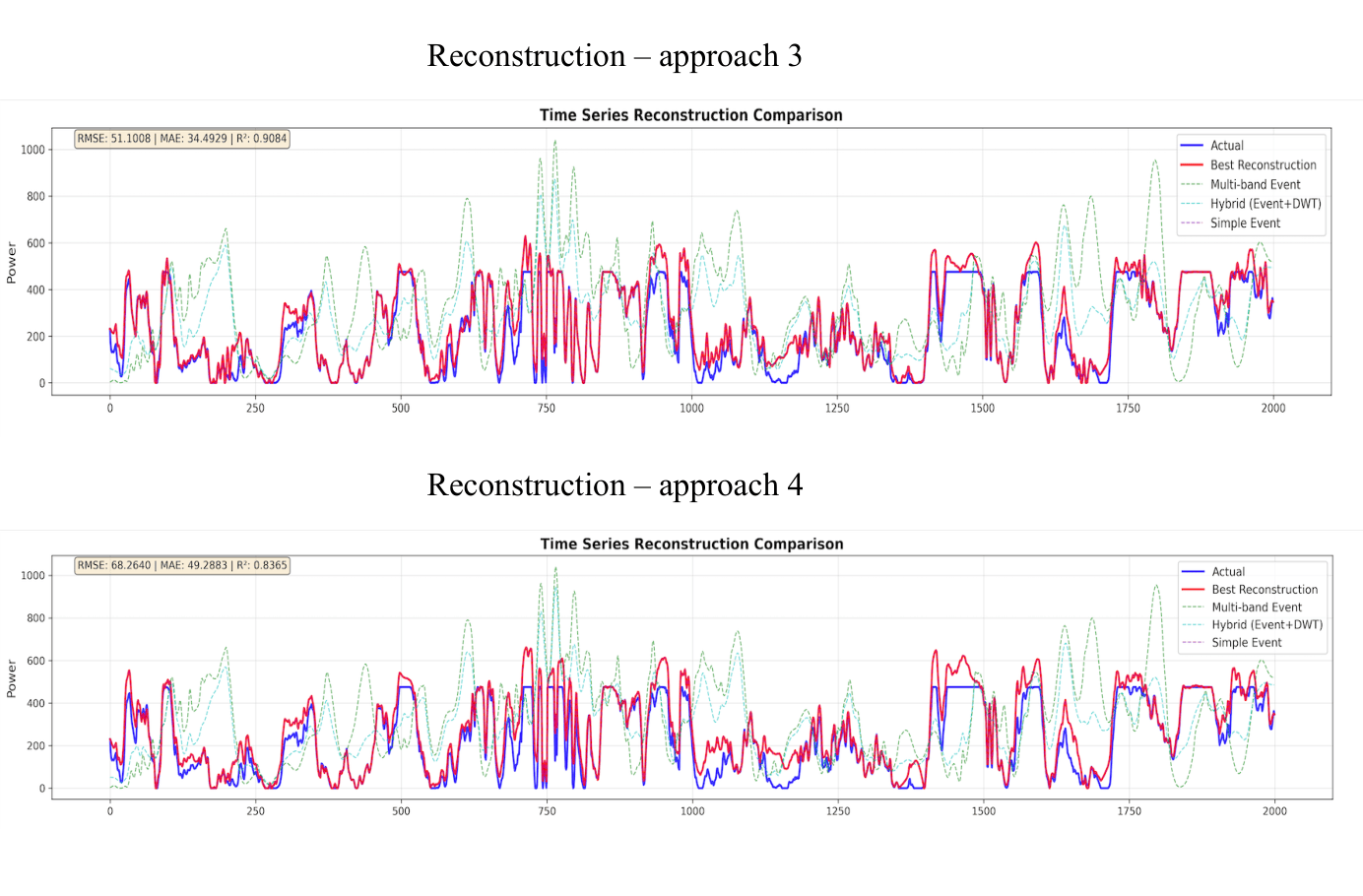}
\caption{Final reconstructed trajectory from predicted events using approach~3 and approach~4}
\label{fig:success-recon-app3}
\end{figure*}

Approach~3 achieves the best reconstruction fidelity,
with $\mathrm{RMSE}\approx51.1$\,MW and $R^2$ of $0.90$ on Baltic Eagle.
Approach~4 follows closely with $R^2\approx0.84$,
producing smoother but slightly attenuated ramps. A comparative reconstruction of both approaches is shown in Figure~\ref{fig:success-recon-app3}. The other methods shown in figure were also applied for comparative analysis where "Simple Event" i.e, our approach turns out to be the best reconstruction method. The other two methods failed because it only uses the poor prediction of Approximation band as base which causes high mismatch. "Simple Event" instead match trends between recent past and prediction and smartly decides the nature of the baseline that actually align closely with the actual Approximation band. These results confirm that event-aware reconstruction is feasible
only when combined with frequency-aware signal anchoring.
Pure event-only reconstruction fails to preserve global trends, while deterministic approximation stabilises long-range structure. It also shows, event aware predictions and then reconstruction can be highly competitive to the traditional time-series forecasting. As then data is shrunk into events, computational cost for complex high volume dataset also gets reduced as discussed in the next point.

\subsubsection{Computational Cost Analysis}
\label{subsubsec:timing}

Operational feasibility of event-aware forecasting
depends critically on training and inference cost.
This subsection evaluates the computational overhead of all four approaches
and clarifies the trade-off between runtime and semantic richness.

\paragraph{Training cost}
Table~\ref{tab:training_times} reports training and total execution times.
Classical approaches are computationally efficient:
SARIMAX-RBA$_\theta$ completes training in $\approx6$ minutes,
while the Random Forest baseline requires $\approx50$ minutes due to its expanded
feature space. Representing the complex time-series in terms of significant and stationary events also helped to compress the data, amplifying the model learning times. This was further proven when SARIMAX was integrated with RandomForest classifier instead of RBA$_\theta$ removing the influence of event-level features. SARIMAX-RF took 3 more minutes to train, which shows incorporating event semantics brings the possibility to improve computational cost associated to complex analytical and forecasting tasks.
Although, the deep event-aware workflows incur substantially higher computational cost compared to first two approaches,
with Approach~3 (LSTM) and Approach~4 (Transformer) requiring over 300 and 400 minutes,
respectively.
This overhead mainly arises from the inherent complex architecture.

\begin{table}[h]
\centering
\caption{Training and total computation times for all forecasting approaches}
\label{tab:training_times}
\begin{tabular}{lcc}
\toprule
\textbf{Approach} & \textbf{Training Time} & \textbf{Total Runtime} \\
\midrule
Approach~1  & 6.04 min & $\approx$8 min \\
Approach~2  & 50.40 min & 55.80 min \\
Approach~3 & 315.70 min & 334.13 min \\
Approach~4 & 399.17 min & 418.33 min \\
\bottomrule
\end{tabular}
\end{table}

\paragraph{Inference cost for zero-shot deployment}
Inference time follows a similar trend.
Table~\ref{tab:zero_shot_34} detail zero-shot transfer runtimes.
The dominant contributors are RBA$_\theta$ event extraction and deep model inference,
particularly for the Transformer due to global attention.
Even so, total inference time remains below 13 minutes per unseen wind farm,
which is acceptable for hourly or day-ahead operational use.

\begin{table}[!t]
\centering
\caption{Zero-shot inference time for Approach~3 and Approach~4.}
\label{tab:zero_shot_34}
\begin{tabular}{lc}
\toprule
\multicolumn{2}{c}{\textbf{Approach~3}} \\
\midrule
\textbf{Component} & \textbf{Time (s)} \\
\midrule
event\_extraction & 175.97 \\
model\_prediction & 315.55 \\
\midrule
\textbf{Total} & \textbf{579.48} \\
\midrule\midrule
\multicolumn{2}{c}{\textbf{Approach~4}} \\
\midrule
\textbf{Component} & \textbf{Time (s)} \\
\midrule
event\_extraction & 179.92 \\
model\_prediction & 485.48 \\
\midrule
\textbf{Total} & \textbf{750.46} \\
\bottomrule
\end{tabular}
\end{table}

While deep workflows are computationally heavier,
they produce substantially richer outputs such as
event occurrence, magnitude, duration, frequency structure,
and physically coherent reconstructions.
This semantic gain justifies the additional cost and supports
deployment in scenarios where decision quality outweighs raw speed.
%________________________
%________________________
\subsection{Ablation Studies}
\label{subsec:ablation}

This section isolates the contribution of event semantics and adaptive feature selection
within the proposed framework. Two complementary ablation levels are considered.
First, a classical baseline ablation evaluates whether explicit RBA$_\theta$ event semantics
provide information beyond pointwise statistical classifiers when applied to SARIMAX
forecasts. Second, component ablations are conducted on the deep event-prediction backbone
(Approach~3) to quantify the individual impact of RBA$_\theta$-derived features and
multi-armed bandit (MAB) feature selection. Since signal reconstruction relies on the DWT
approximation band, no DWT-removal ablation is performed.

\subsubsection{Baseline Ablation: SARIMAX with and without RBA$_\theta$ Information}

Two variants are compared that are a post-hoc RBA$_\theta$ event extraction pipeline
(Approach~1) and a similar SARIMAX integrated to a RandomForest Classifier for event classification removing the influence of RBA$_\theta$ event semantics. While both approaches yield comparable trajectory accuracy, their event-prediction capability differ markedly.

In power forecasting, Approach~1 achieves a lower mean absolute error
(15.36~kW versus 16.57~kW), whereas SARIMAX--RF attains a lower RMSE
(18.04~kW versus 23.03~kW) and a slightly higher coefficient of determination
($R^2=0.9896$ versus $0.9830$). 
Despite similar trajectory fidelity, event prediction performance diverges strongly.
Post-hoc RBA$_\theta$ extraction achieves an F1-score of $0.91$, corresponding to a relative improvement of $+16.8\%$ compared to SARIMAX--RF.
Precision and recall increase by approximately $+14.2\%$ and $+19.5\%$, respectively.
This gap demonstrates that the temporal ramp grammar encoded by RBA$_\theta$ cannot be
recovered reliably from pointwise statistical features alone. The SARIMAX--RF pipeline
remains competitive primarily for stationary regimes (F1 $\approx 0.93$), where temporal
structure is weak and event boundaries are less informative.

\subsubsection{Deep-Model Ablations: Event Semantics and Feature Selection}

Component ablations are next conducted on the deep event-prediction backbone to quantify
the role of semantic inputs and adaptive feature selection. We started with a vanilla LSTM baseline that uses
ground-truth event labels for supervision but does not ingest RBA$_\theta$-derived features.
An event-aware variant augments the same backbone with RBA$_\theta$ inputs. Additional
ablations remove RBA$_\theta$ features or disable MAB selection within Approach~3.
Table~\ref{tab:ablation_summary} summarises the quantitative impact of these ablations,
focusing on precision and F1 as operationally relevant metrics. For Approach~3 component
ablations, results are reported at 1\,h and 24\,h horizons to contrast short-horizon
noise sensitivity against long-horizon temporal context.

\begin{table}[t]
\centering
\caption{Ablation summary for approach~3 in terms of precision and F1}
\label{tab:ablation_summary}
\small
\setlength{\tabcolsep}{6pt}
\begin{tabular}{lcc}
\toprule
\textbf{Setting} & \textbf{Precision} & \textbf{F1} \\
\midrule
Vanilla LSTM (no RBA inputs) & 0.9764 & 0.9787 \\
RBA$_\theta$-augmented LSTM  & 0.9751 & 0.9793 \\
\midrule
Approach~3 without MAB (1\,h) & $\approx0.26$ & 0.39 \\
Approach~3 without RBA features (1\,h) & $\approx0.06$ & 0.09 \\
Approach~3 without RBA features (24\,h) & $\approx 0.64$ & 0.64 \\
Approach~3 without MAB (24\,h) & $\approx0.81$ & 0.75 \\
\bottomrule
\end{tabular}
\end{table}

Augmenting a strong LSTM backbone with RBA$_\theta$ features yields only a marginal absolute
gain in F1 ($+0.06\%$), reflecting that the vanilla model already fits the labels well.
However, this comparison is performed under a deliberately coarse binary formulation,
where the task is limited to predicting whether an event occurs or not, without resolving
event type, magnitude, or temporal structure. Under this setting, predictions are
inherently noisy and information-poor, which constrains the observable benefit of semantic
augmentation. The consistent improvement nonetheless indicates that RBA$_\theta$ features
act as a semantic regulariser rather than a capacity expansion mechanism, with their
primary advantage emerging in structured, multi-class and horizon-aware event settings
rather than in binary detection alone.

In contrast, removing RBA$_\theta$ features from the Approach~3 backbone causes a severe
performance collapse at short horizons, with F1 dropping to $0.09$ at 1\,h.
Even with increased temporal context at 24\,h, performance recovers only partially
(F1 and precision $\approx 0.64$), demonstrating that explicit event semantics are
essential for stabilising ramp discrimination under short-term turbulence.

Disabling MAB feature selection also degrades performance, though less
catastrophically than removing semantic inputs. At 1\,h, F1 drops to $0.39$, consistent
with high-dimensional feature noise overwhelming the learner. At 24\,h, performance
recovers F1 to $0.75$, but remains inferior to the full model, indicating residual
sensitivity to feature redundancy. While mutual-information-based selection avoids
complete collapse, it does not match the robustness of adaptive, horizon-aware MAB
selection.
Overall, the ablation results establish a clear hierarchy of contributions. RBA$_\theta$
features provide the dominant gain in robustness at short horizons, where false alarms and
boundary fragmentation are most likely. MAB feature selection is critical for preventing
feature overload and preserving precision-oriented behaviour as the candidate feature pool
exceeds approximately $1{,}400$ dimensions. Reliable event-aware forecasting therefore
requires both explicit semantic inputs and adaptive feature selection to remain stable
under operational noise and regime variability.

% ======================================================
\subsection{Zero-Shot Transfer Across Wind Farms}
\label{subsec:transfer_results}

Generalisation is evaluated by applying Baltic Eagle–trained models
to Baie de Saint--Brieuc and London Array without retraining.
Approach~4 achieves:
\[
\mathrm{F1}_{24\mathrm{h}}\approx0.62,\quad
\mathrm{Precision}_{24\mathrm{h}}\approx0.60
\]
on Baie de Saint--Brieuc,
with successful trajectory reconstruction from the predictions. Similar performance is achieved for the other dataset as well which proves the consistency in performance.

Approach~3 yields comparable reconstruction accuracy
with 5\% higher precision but with the same (5\%) lower F1-score. Both approaches repetitively produced around 0.60 precision in event prediction on completely unseen data, that too completely without fine-tuning.
These results demonstrate that
DWT representations and RBA$_\theta$ event semantics
encode somewhat site-invariant structure that transfers across geophysical regimes.
Across all experiments, three consistent findings emerge.
First, event predictability increases with horizon as temporal context aligns with
physical ramp formation scales.
Second, magnitude and duration information concentrates in mid-frequency bands,
validating the frequency-aware design.
Third, reconstruction fidelity and transferability confirm that event-first representations provide a robust abstraction
for operational wind-power forecasting.
Together, these results establish the proposed framework
as a practical and generalisable alternative to trajectory-first forecasting,
offering improved semantic clarity, robustness, and cross-site applicability.

\subsection{Agentic Workflow Selection}
\label{subsec:agentic-workflow}

The agentic system was validated on Baltic Eagle wind farm data (350,640 samples, 40-year ERA5 reanalysis). Bootstrap initialization from 8 expert rules provided 72 synthetic priors. On the first real execution, the agent detected local optimum (Approach/Workflow~3 over-represented at 38.9\%) and forced exploration of Workflow~1(SARIMAX-RBA). Despite forced intervention, Workflow~1
achieved $R^2$ of 0.98, MAE of 15.57, and F1 of 0.91, establishing a reward of 0.820. With adaptive exploration and contradiction detection, the system achieves more and more accurate workflow selection capability after multiple executions, with cumulative reward improvement over fixed rules. Stratified sampling enables sub-3-second data characterization even for large datasets while maintaining minimal feature extraction error, making the system suitable for real-time deployment.

\section{Limitations}
\label{sec:limitation}

Despite the strong empirical performance of the proposed event-aware forecasting
framework, several limitations remain.
First, unlike LSTM-based Approach~3, horizon-specific
Transformer models required several hours of training, and removing core components
such as RBA$_\theta$ inputs or MAB feature selection for ablation frequently destabilised attention
and led to non-convergent training. As a result, the contribution of individual
components in Approach~4 could not be isolated.
Second, short-horizon event prediction (1-3~h) still remains challenging. Although adaptive
feature selection and frequency-domain representations improve performance over
classical baselines, precision and $R^2$ are substantially lower than at longer
horizons. This limitation might be due to the absence of numerical weather prediction inputs, as short-term wind ramps are often driven by mesoscale phenomena
not inferable from historical data alone.

Third, generalisation across heterogeneous domains has not yet been established.
Zero-shot application to a different dataset was not feasible mostly due to fundamental feature mismatch and domain-specific data constraints. Remodification of the workflows in context of those datasets were also performed and currently the multivariate prediction part is highly dependent on input data quality and balance across variables. When individual entities (e.g., price zones
or turbines) exhibit strong imbalance, sparsity, or inconsistent dynamics, model
performance degrades substantially. In contrast, when data quality is high and sufficiently long temporal context is available for training, the model exhibits strong performance. The current design therefore favours data-rich settings and is not yet suitable for short or highly imbalanced datasets, limiting its applicability
across heterogeneous domains.

Fourth, explicit uncertainty quantification could not be incorporated due to computational constraints. Bayesian techniques such as Monte Carlo dropout, deep ensembles, and
probabilistic inference were incompatible with the memory and training requirements of the multi-band LSTM and Transformer models. While RBA$_\theta$ and Hawkes Causality implicitly captures event-level uncertainties through persistence and boundary variability, calibrated probabilistic estimates are not provided.

Lastly, the event extraction stage relies on enhanced RBA$_\theta$-Traditional thresholding method which is statistical based. Bayesian adaptive thresholding  via RBA$_\theta$ RF-MCMC was omitted due to its high computational cost.
The evaluation is constrained by dataset and modelling scope. Experiments
primarily rely on reanalysis or synthetic data and do not capture operational SCADA
effects such as curtailment or sensor faults. Moreover, the framework models a single
wind farm at a time and does not account for spatial interactions, limiting
applicability in multi-farm settings.
These limitations arise from computational constraints, data availability, and the
intrinsic complexity of atmospheric dynamics, and they define clear directions for
future work.

\section{Conclusion and Future Scope}
\label{sec:conclusion-future}

This work advances wind-power forecasting by reframing it as an
\emph{event-centric, frequency-aware, and agentically coordinated} problem rather
than a dense trajectory regression task. Motivated by the operational reality that
power-system decisions depend on the timing, magnitude, duration, and persistence
of events, the proposed framework aligns forecasting objectives with decision-making
needs. Empirical results across multiple workflows demonstrate that wind-power ramps
are inherently multi-scale phenomena, with predictive information concentrated in
specific mid-frequency bands. Embedding enhanced RBA$_\theta$ event semantics,
multi-resolution signal decomposition, and adaptive feature selection directly into
the learning pipeline enables stable event prediction and physically coherent
reconstruction. Reconstruction achieves competitive
trajectory accuracy ($R^2 \approx 0.90$) while ablation studies confirm that event
semantics and adaptive feature selection are structural requirements rather than
auxiliary enhancements. Comparative analysis further shows that LSTM- and
Transformer-based workflows offer complementary strengths, motivating an agentic
forecasting paradigm in which specialised models are dynamically selected based on
horizon, uncertainty, and data characteristics rather than enforcing a single
monolithic solution.

Future work extends naturally from these findings along three primary directions.
First, incorporating spatial and graph-based models can enable explicit modelling
of spatial correlations and event propagation across turbines, wind farms, or
market zones, capturing inter-location autocorrelation effects that are not
addressed in the current single-site formulation. Second, formalising multivariate
event prediction remains an important extension, including principled treatment
of variable coupling, data imbalance across entities, and context requirements
for reliable learning in high-dimensional settings. Third, advancing real-time
adaptability under uncertainty is essential for operational deployment. This
includes developing lightweight uncertainty-aware mechanisms and adaptive control
strategies that allow event-first models to remain stable under streaming data,
regime shifts, and rapidly evolving system conditions. Together, these directions
aim to generalise the proposed framework beyond wind power and enable scalable,
interpretable, and robust forecasting systems for complex energy applications.

\end{document}